\documentclass[lettersize,journal]{IEEEtran}

\ifCLASSOPTIONcompsoc
\usepackage[nocompress]{cite}
\else
\usepackage{cite}
\fi

\ifCLASSINFOpdf
\else
\fi

\usepackage{gensymb}
\usepackage[table]{xcolor}
\definecolor{Gray}{gray}{0.9}
\usepackage{amsmath,amssymb,amsfonts}
\usepackage{graphicx}
\usepackage{textcomp}
\usepackage{tikz}
\usetikzlibrary{bayesnet}

\usepackage{mathrsfs}
\usepackage[colorlinks,linkcolor=red,citecolor=blue]{hyperref}
\usepackage{amsmath}
\usepackage{amsfonts,subfigure}
\usepackage{amssymb}
\usepackage{multirow}
\usepackage{bm}
\usepackage{sansmath}
\usepackage{booktabs}
\usepackage[ruled]{algorithm2e}

\usepackage{graphicx}

\usepackage{pifont} 
\usepackage{amsthm}

\usepackage{tabularx, makecell, multirow} 
\graphicspath{{figure/}}

\usepackage{bm}
\usepackage{color,soul}

\usepackage{tikz}

\usepackage{multirow}
\usepackage{adjustbox}
\usepackage{environ}
\usepackage{setspace}

\usepackage{enumitem}

\newcommand{\secref}[1]{Sec. \ref{#1}}
\newcommand{\tableref}[1]{Table~\ref{#1}}
  
\newcommand{\figref}[1]{Figure~\ref{#1}} 
\graphicspath{{image/}}

\usepackage{makecell}
\definecolor{Gray}{gray}{0.9}
\usepackage{graphicx}

\begin{document}
	
	\title{Evaluating the Generalization Ability of Spatiotemporal Model in Urban Scenario }
	
	%
	%
	%
	%
	

	\author{Hongjun~Wang,
		Jiyuan~Chen,
		Tong~Pan,
		Zheng~Dong,\\
		Lingyu~Zhang,
		Renhe~Jiang,
		and
		Xuan Song
		\IEEEcompsocitemizethanks{
			\IEEEcompsocthanksitem Hongjun Wang, Jiyuan Chen,  Tong Pan,  Zheng Dong  and Lingyu Zhang are with (1) SUSTech-UTokyo Joint Research Center on Super Smart City, Department of Computer Science and Engineering
			(2) Research Institute of Trustworthy Autonomous Systems, Southern University of Science and Technology (SUSTech), Shenzhen, China.
			E-mail: {wanghj2020,11811810}@mail.sustech.edu.cn, pant@sustech.edu.cn,  zhengdong00@outlook.com, and zhanglingyu@didiglobal.com. 
			\IEEEcompsocthanksitem Xuan Song is with (1) School of Artificial Intelligence, Jilin University  (2) Research Institute of Trustworthy Autonomous Systems, Southern University of Science and Technology (SUSTech), Shenzhen, China. Email: songxuan@jlu.edu.cn.
			 \IEEEcompsocthanksitem R. Jiang is with Center for Spatial Information Science, University of
			 Tokyo, Tokyo, Japan. Email: jiangrh@csis.u-tokyo.ac.jp.
			\IEEEcompsocthanksitem Corresponding to  Xuan Song;
		}
	}

	%
	%

\markboth{Journal of \LaTeX\ Class Files,~Vol.~XX, No.~X, August~201X}%
{Shell \MakeLowercase{\textit{et al.}}: Bare Demo of IEEEtran.cls for Computer Society Journals}
%

\maketitle
\begin{abstract}
Spatiotemporal neural networks have shown great promise in urban scenarios by effectively capturing temporal and spatial correlations. However, urban environments are constantly evolving, and current model evaluations are often limited to traffic scenarios and use data mainly collected only a few weeks after training period to evaluate model performance. The generalization ability of these models remains largely unexplored. To address this, we propose a Spatiotemporal Out-of-Distribution (ST-OOD) benchmark, which comprises six urban scenario: bike-sharing, 311 services, pedestrian counts, traffic speed, traffic flow, ride-hailing demand, and bike-sharing, each with in-distribution (same year) and out-of-distribution (next years) settings. We extensively evaluate state-of-the-art spatiotemporal models and find that their performance degrades significantly in out-of-distribution settings, with most models performing even worse than a simple Multi-Layer Perceptron (MLP). Our findings suggest that current leading methods tend to over-rely on parameters to overfit training data, which may lead to good performance on in-distribution data but often results in poor generalization. We also investigated whether dropout could mitigate the negative effects of overfitting. Our results showed that a slight dropout rate could significantly improve generalization performance on most datasets, with minimal impact on in-distribution performance. However, balancing in-distribution and out-of-distribution performance remains a challenging problem. We hope that the proposed benchmark will encourage further research on this critical issue.  \textcolor{magenta}{Codes are available at \textcolor{black}{\href{https://github.com/Dreamzz5/ST-OOD}{GitHub}}}
	\end{abstract}
	
	\begin{IEEEkeywords}
		Traffic Forecasting, Urban Computing, Domain Generalization
\end{IEEEkeywords}

\section{Introduction}

\IEEEPARstart{O}{ver} a decade ago, research had already demonstrated that machine learning systems could exhibit significant failures when evaluated on data outside the domain of their training examples, primarily due to their heavy dependence on the training distribution \cite{torralba2011unbiased}. For instance, self-driving car systems have been shown to perform poorly under conditions that deviate from their training environments, such as variations in lighting \cite{dai2018dark}, weather \cite{volk2019towards}, and object orientations \cite{alcorn2019strike}.

\begin{figure}
	\centering
	\includegraphics[width=1 \linewidth]{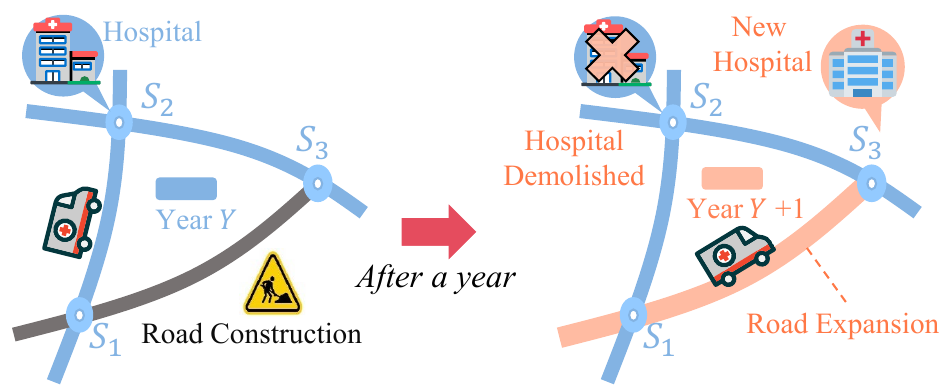}
	\caption{One example of a spatiotemporal shift is the dynamic evolution of urban areas. As new roads or points of interest continuously emerge and older ones are gradually removed, new traffic demands and spatiotemporal relationships are created. However, current spatiotemporal models have not yet been tested under such evolving conditions.} 
	\label{fig:motivation}
\end{figure}

Despite the significant model degradation observed in various domains, there is limited research addressing this issue within the urban-related applications. Rapid urbanization in recent years has presented substantial challenges to accurate traffic forecasting~\cite{zhang2020curb}. The continuous growth and development of urban areas have led to significant changes in traffic patterns and mobility demands, imposing rigorous requirements on the generalization capabilities of these models. An illustrative example of Spatiotemporal Out-of-Distribution  is shown in \figref{fig:motivation}. Prior to year $Y$, a road connected $S_1$ and $S_3$ is under construction, while $S_2$ housed a hospital. Consequently, a strong spatial correlation existed between $S_1$ and $S_2$ relative to $S_3$. However, a significant shift occurred in the subsequent year: the hospital in $S_2$ was demolished, and a new healthcare facility was constructed in $S_3$, introducing a spatial out-of-distribution scenario. To compound this, the road linking $S_1$ and $S_3$ underwent expansion, creating a temporal OOD condition. These transformative changes engendered novel spatiotemporal relationships, prompting an inquiry into the adaptability of the ST-model to such evolving circumstances.

Despite the proliferation of spatiotemporal datasets, many are constrained to evaluating model performance over relatively short time horizons, such as METR-LA \cite{li2018diffusion} and PEMS  \cite{yu2018spatio}. While LargeST \cite{liu2024largest} has recently emerged as a dataset encompassing several years of data, it is exclusively focused on traffic flow scenarios. Consequently, a systematic evaluation of the generalization performance of existing models across diverse urban scenarios remains an unmet need. To address this issue, we have collected six categories of urban data, including shared bicycles, 311 service, pedestrians, traffic speed, traffic flow, bike-sharing, and ride-hailing demand.  We have designed both in-distribution and out-of-distribution testing scenarios. Specifically, models are trained on data from year $A$ and then tested on data from both year $A$ and year $A+1$. By comparing the performance in year $A$ with that in year $A+1$, we can more effectively evaluate the generalization capabilities of the models.

Our research reveals that while mainstream approaches, such as STAEformer \cite{liu2023spatio} and ARCRN \cite{bai2020adaptive}, aim to transcend the limitations of real-world topological graphs through purely model-driven techniques, their generalization performance often lags despite strong results on in-distribution data. This shortfall is largely attributable to the model's tendency to capture spurious relationships \cite{pearl2009causality}, which tend to be unstable in practical, real-world applications.
In OOD scenarios, Tobler's First Law of Geography \cite{tobler1970computer}—stating that everything is related to everything else, but near things are more related than distant things—exhibit a more robust and reliable influence. Methods that integrate real-world topological graphs with adaptive graphs, such as GWNet \cite{wu2019graph} and STGODE \cite{fang2021spatial}, demonstrate markedly better generalization performance in OOD scenarios.
Interestingly, while discarding real-world topological graphs may seem like a radical departure, STID \cite{shao2022spatial} illustrates that reducing the model size and relying exclusively on node embeddings can mitigate the risk of the model learning spurious relationships, as fewer parameters are involved in the learning process \cite{asgari2022masktune}.

While recent efforts \cite{xia2024deciphering, zhou2023maintaining, wang2024stone} have sought to tackle the spatiotemporal OOD challenge, the lack of a dedicated, comprehensive dataset for robust evaluation remains a critical gap in the existing research. Current methods based on spatiotemporal OOD, either spanning only a few weeks \cite{xia2024deciphering} or relying on artificially generated OOD scenarios \cite{zhou2023maintaining,wang2024stone}, have yet to undergo comprehensive evaluation.  After evaluating these methods on our proposed benchmark, we found that mainstream spatiotemporal OOD methods generally achieve better generalization by underfitting the spatiotemporal model, leading to inferior performance on in-distribution data. Moreover, their performance is inconsistent, with many scenarios even underperforming STID.

Our contributions are summarized as follows:
\begin{itemize}[leftmargin=*]
	\item[\ding{117}] We propose ST-OOD, the first open-source spatiotemporal benchmark specifically designed to evaluate the generalization capabilities of spatiotemporal models in OOD scenarios. As a pioneering study, we envision that our results and findings will present exciting opportunities for advancing methods in spatiotemporal generalization.
	\item[\ding{117}] Extensive research emphasizes that mitigating the negative impact of inductive biases is crucial for determining a model's generalization performance. For example, avoiding over-reliance on graph information is essential to prevent learning spurious correlations.
	\item[\ding{117}] We further evaluated current approaches for spatiotemporal OOD scenarios and found that they perform worse than even simple models like STID or MLP.  Further research is needed to identify architectural designs that can enhance the robustness of models in spatiotemporal scenario.
\end{itemize}

\section{Related Work}

\subsection{Spatiotemporal Benchmarks}
The spatiotemporal research community has benefited from a diverse array of public datasets and benchmarks across various urban computing domains. In transportation, datasets like METR-LA \cite{li2018diffusion} and PEMS series \cite{guo2019attention} offer rich spatio-temporal data for traffic prediction tasks, while datasets such as NYC taxi \cite{sun2020predicting} and bike-sharing records \cite{chai2018bike} enable research on urban mobility patterns. Environmental studies are supported by air quality datasets from major cities \cite{lin2018exploiting} and meteorological data collections like WeatherBench \cite{lin2022conditional}. Public safety research utilizes crime datasets from metropolitan areas \cite{xia2021spatial} and natural disaster records \cite{yang2022spatio}. In the public health domain, datasets tracking the spread of diseases, including COVID-19 \cite{wang2022causalgnn} and influenza \cite{deng2019graph}, are available. Complementing these datasets, domain-specific benchmarks such as BasicTS \cite{shao2023exploring} for traffic prediction and WeatherBench2  \cite{rasp2023weatherbench} for meteorological forecasting provide standardized evaluation frameworks. This ecosystem of open resources fosters reproducibility, enables fair model comparisons, and drives innovation in spatiotemporal research across diverse urban computing applications. 

Although numerous datasets have been introduced, the generalization capabilities of spatiotemporal models remain inadequately understood. A major concern stems from the fact that conventional datasets typically evaluate models only a few weeks after training, providing limited insight into their long-term robustness. To address this limitation, this paper introduces a systematic benchmark, ST-OOD, marking the first comprehensive effort to construct a spatiotemporal out-of-distribution  scenario for extensive validation of existing models.

\subsection{Spatiotemporal Models}

In recent years, deep neural networks, particularly graph neural networks (GNNs), have become the preferred approach for spatiotemporal prediction \cite{li2018diffusion,cao2020spectral,wang2023easy,wang2023multi,wang2022st}. Researchers typically combine GNNs to capture complex spatial dependencies with recurrent neural networks (RNNs) or temporal convolutional networks (TCNs) to model temporal dependencies. For instance, the DCRNN \cite{li2018diffusion} initially introduced a novel diffusion convolution that works alongside GRU. Following this, a series of related works emerged, such as AGCRN \cite{bai2020adaptive}, and DGCRN \cite{li2021dynamic}, which also employed RNNs and their variants. 
To optimize RNNs for faster training and leverage parallel computing, numerous methods such as STGCN \cite{yu2018spatio}, GWNet \cite{wu2019graph}, and DMSTGCN \cite{han2021dynamic} replaced RNNs with dilated causal convolutions for temporal pattern modeling. Other approaches focus on graph structures; for example, ASTGCN \cite{guo2019attention} uses attention mechanisms to model both spatial and temporal correlations. Recently, there has been a trend towards coupling GNNs with neural ordinary differential equations to generate continuous layers for modeling long-range spatiotemporal dependencies, as seen in STGODE \cite{fang2021spatial} and STGNCDE \cite{choi2022graph}. Another trend involves using dynamic graphs to construct adjacency matrices at different time steps to capture dynamic correlations between nodes, exemplified by DGCRN \cite{li2021dynamic}, DSTAGNN \cite{lan2022dstagnn}, and D$^2$STGNN \cite{shao2022decoupled}. 

\subsection{Spatiotemporal OOD Models}
Earlier research in time series analysis has already explored OOD issues ~\cite{du2021adarnn, yao2022wild, deng2023spatio,wang2024stone}. Recently, OOD phenomena have begun to attract attention in the field of spatiotemporal urban prediction. As cities evolve and people's mobility preferences change over time, the temporal and spatial contexts of training and testing data can easily diverge. In the latest studies by \cite{xia2024deciphering, zhou2023maintaining}, researchers have attempted to address OOD scenarios through causal inference in existing spatiotemporal prediction processes. Specifically, Xia et al. \cite{xia2024deciphering} employs disentanglement blocks for backdoor adjustment, separating temporal environments from input data and utilizing frontdoor adjustment with edge-level convolution to model the causal chain effect. Meanwhile, Zhou et al. \cite{zhou2023maintaining} transforms invariant learning into capturing stable, trainable weights. However, the aforementioned methods either rely on comparisons using existing datasets or manually create OOD conditions.
However, the generalization capabilities of traditional models remain unclear. Concerns arise from the fact that traditional datasets typically test models only a few weeks after training. To address this gap, this paper introduces a systematic benchmark named ST-OOD, the first attempt to build a comprehensive spatiotemporal OOD scenario to extensively validate existing models.

\section{Problem Formulation}
Let $\mathcal{G} = (V, E, A)$ represent a spatial network (e.g., road network, sensor network, grid, etc.), where $V$ and $E$ denote the sets of vertices and edges, respectively. We use $A$ to represent the adjacency matrix of the spatial network $\mathcal{G}$. We further define the graph signal matrix $X_{(t)} \in \mathbb{R}^{N \times C}$ for $\mathcal{G}$, where $C$ represents the dimensionality of the features, $n = |V|$ is the number of vertices, and $X_{(t)}$ denotes the observations of the spatial network $\mathcal{G}$ at time step $t$. 
Overall, the spatiotemporal task aims to learn a multi-step prediction function $f$ based on past observations:
\begin{equation}
	\begin{aligned}
		f((X_{(t-\alpha)}, \ldots,X_{t-1}), \mathcal{G}) \rightarrow (X_{(t)}, X_{(t+1)},\ldots,X_{t+\beta})
	\end{aligned}
\end{equation}
where $\alpha$ is the input length of past time step observations, and $\beta$ is the number of future steps we wish to predict.

\begin{table*}
	\caption{Summary of proposed benchmark. ST-OOD validates a model's generalization ability by comparing its performance on data from the same year and different years.}\label{tab:inf}
	\renewcommand{\arraystretch}{1.1}
	\resizebox{\textwidth}{!}{
		\begin{tabular}{cccccccc}
			\toprule Application & City/State &Train Year &Train Span  &Test Year& Test Span  & \# Units  & Interval \\
			\hline Bike-sharing & Chicago& 2019&$01 .01-10 .19$  & 2019/2020 & $10 .20-12 .31$  & 609 & 60 mins \\
			Ride-sharing & Chicago & 2013 & $01 .01-10 .19$ & 2013/2014 & $10 .20-12 .31$  & 77 & 60 mins \\
			Vehicle Speed & NYC & 2019 & $03 .01-05 .12$  & 2019/2020 & $05 .13-05 .31$& 139 & 5 mins \\
			311 Service & NYC & 2019 & $01 .01-10 .19$ & 2019/2020 & $10 .20-12 .31$  & 71 & 60 mins \\
			Pedestrian  & Zurich & 2019 & $01 .01-10 .19$ & 2019/2020 & $10 .20-12 .31$ & 99 & 60 mins \\
			Traffic Flow   &  California & 2016 & $07 .01-08 .18$ & 2017/2018 &  $08 .19-08 .31$  & 170 & 5 mins \\
			
			\bottomrule
		\end{tabular}
	}
\end{table*}

Current evaluation methods \cite{wu2019graph, bai2020adaptive, wu2020connecting} typically assume a consistent graph relationship within the dynamic feature matrix $X_{(t)}$, expecting trained models to perform well in the short-term future (e.g., over weeks or months). While this may seem reasonable for current spatiotemporal benchmarks where graph relationships remain stable over a limited span, it raises concerns about the performance of existing spatiotemporal networks in long-term scenarios (e.g., a year later). Our work explores a more practical scenario, where the graph relationships during training and testing may differ, and the training graph relationships may evolve over time.  Although some recent studies on spatiotemporal OOD have been proposed \cite{xia2024deciphering, zhou2023maintaining}, such as attempts to address the spatiotemporal OOD problem using causal inference methods \cite{pearl2009causality}, no substantial dataset has yet been introduced. These studies typically focus on scenarios spanning only a few weeks \cite{xia2024deciphering} or manually create OOD conditions \cite{zhou2023maintaining}.

\section{The ST-OOD Benchmark}
\begin{figure}[t]
	\centering
	\includegraphics[width=1 \linewidth]{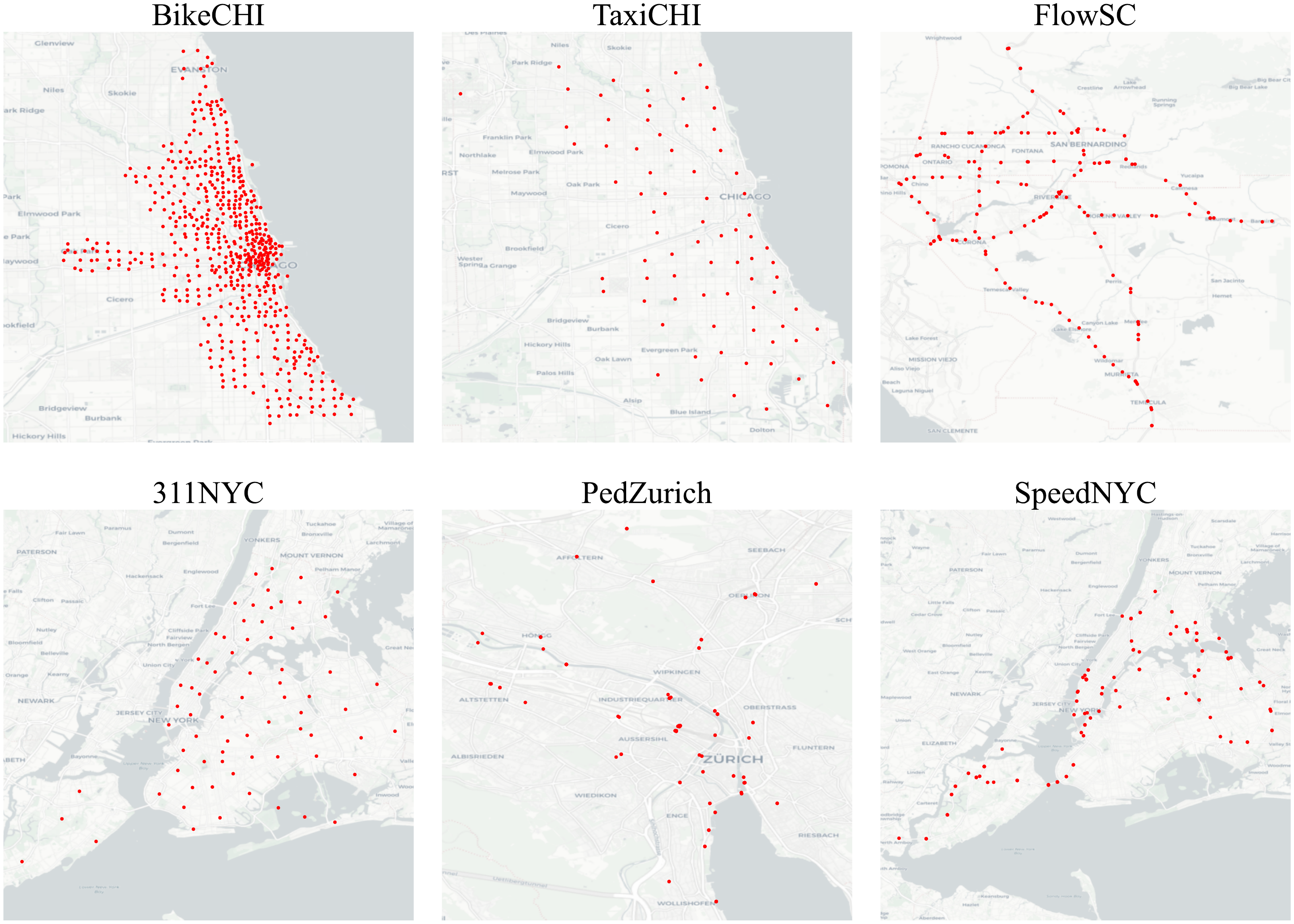}
	\caption{Map visualization of node distribution.} 
	\label{fig:geoplot}
\end{figure}

This section formally introduces the proposed ST-OOD benchmark, designed to comprehensively evaluate the spatiotemporal generalization capabilities of current models across various scenarios. We begin by detailing the collection and organization of ST-OOD in \secref{sec:1}. The detailed information is listed on \tableref{tab:inf} and the node distribution is visualized in \figref{fig:geoplot}. Then, in \secref{sec:2}, we conduct an in-depth data analysis to gain deeper insights into ST-OOD.

\subsection{Data Collection and Preprocessing}\label{sec:1}
In the following section, we provide a detailed description of the data collection process. All datasets were uniformly trained using data from the previous year, while testing was conducted on the same dates for both the same year and the following year to evaluate model generalization ability. We adopted the data partitioning strategy established in prior work \cite{jiang2021dl}, which chronologically divides the data into training, validation, and testing subsets with a 6:2:2 ratio. We have devised both in-distribution and out-of-distribution testing protocols to rigorously assess model performance. Specifically, models are trained on data from year $A$ and subsequently evaluated on datasets from both year $A$ (in-distribution) and year $A+1$ (out-of-distribution). By juxtaposing the model's performance on year $A$ data against its performance on year $A+1$ data, we can more comprehensively gauge the model's generalization capabilities and robustness to temporal shifts in data distribution. Additionally, for each dataset, we generated adjacency matrices based on the haversine (great-circle) distance between units. Units within 500 meters were considered connected (denoted as 1), while those beyond this threshold were considered disconnected (denoted as 0). For the 311 service dataset, where the statistical units are community districts, the adjacency matrix was constructed based on the geographic topology and connectivity of these districts.

\begin{itemize}
	\item[\ding{117}] \textbf{Pedestrians Data.} We used pedestrian counting data collected by the Civil Engineering Office of the City of Zurich\footnote{\url{http://www.stadt-zuerich.ch/tiefbauamt}} through automated counting stations distributed across the city. This dataset includes location information for approximately 99 counting points throughout Zurich. The data is aggregated at 15-minute intervals, which we further consolidated into 1-hour intervals and selected two years of data, spanning from January 1, 2019, to December 1, 2020.
	
	\item[\ding{117}] \textbf{Taxi Demand Data.} The taxi dataset was collected from the Chicago Open Data Portal\footnote{\url{https://data.cityofchicago.org/}}. It includes information on taxi trips such as pickup and drop-off times, locations, and fares. In our dataset, we focus solely on the pickup and drop-off information for each record. We aggregated the hourly pickup records for 77 community areas, with the data spanning from January 1, 2013, to December 31, 2014.
	
	\item[\ding{117}] \textbf{Bike-sharing Data.} The shared bicycle dataset is sourced from the Chicago Open Data Portal (CHI, Divvy Bikes System Data\footnote{\url{https://www.divvybikes.com/system-Data.}}). Each record contains information such as the start station, start time, end station, and end time. In this paper, we focus solely on the starting information for each record. The dataset includes a total of 585 stations and spans from January 1, 2013, to December 31, 2014. 
	
	\item[\ding{117}] \textbf{Traffic Speed Data.} We collected real-time traffic speed data from the New York City Department of Transportation's Traffic Management Center (TMC)\footnote{\url{https://www.nyc.gov/html/dot/html/motorist/atis.shtml}}. The real-time traffic information is recorded from 135 sensor sprimarily located on major arterials and highways across the New York City. Our experiment focused on the period from March 1 2019  to May 31 2019 for training and March 1 2020  to May 31 2020.
	
	\item[\ding{117}] \textbf{Traffic Flow Data.} We collected real-time traffic flow data from the California Department of Transportation  (CalTrans) Performance Measurement System (PEMS)\footnote{\url{https://pems.dot.ca.gov/}}, an online platform delivering real-time traffic data sourced from 18,954 loop detectors strategically positioned across the California state highway system.
	PEMS08 covers Riverside and San Bernardino Counties in Southern California. Our experiment focused on the period from July 1 2016 to August 31 2016 for training and July 1 2017 to August 31 2017. 
	
	\item[\ding{117}] \textbf{311 Service Data.} 311 service requests cover all non-emergency requests in the city, including but not limited to noise complaints, air quality issues, and reports of unsanitary conditions. The 311 calls for New York City (NYC) are publicly available, and the dataset can be accessed at NYC 311 Service Requests from 2010 to Present\footnote{\url{https://data.cityofnewyork.us/}}. We used NYC community districts to aggregate the number of requests per hour in each area and constructed the adjacency matrix based on geographical connectivity.
\end{itemize}

\begin{figure*}[t]
	\centering
	\subfigure[311 NYC]{\includegraphics[width=0.48\linewidth]{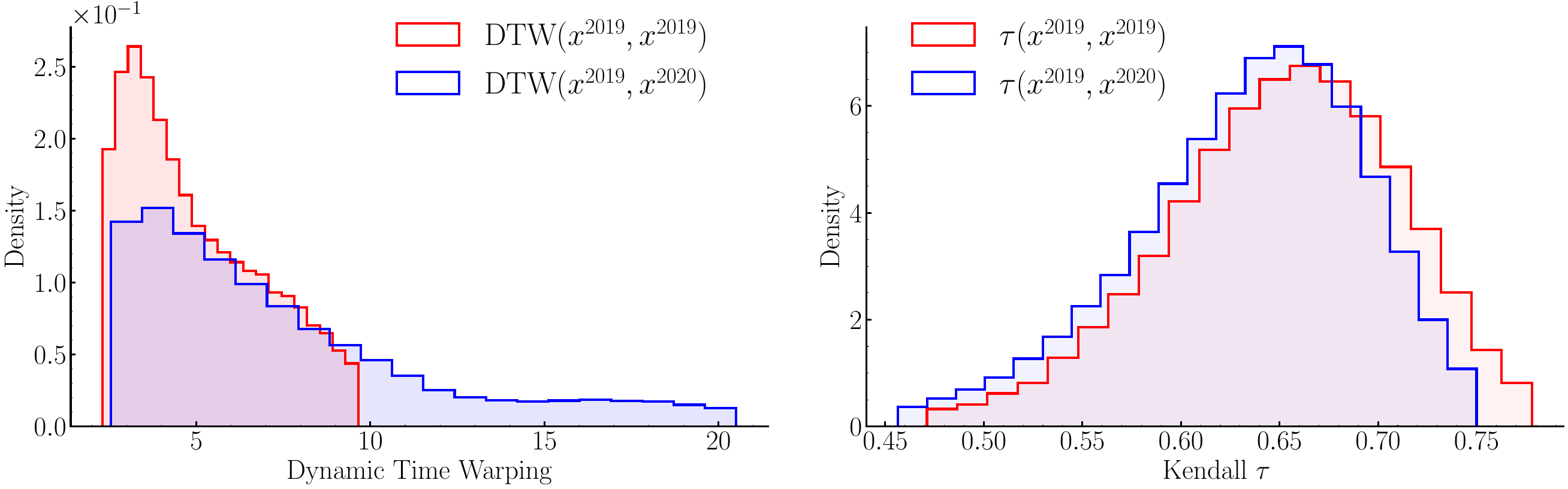}}
	\subfigure[Flow SC]{\includegraphics[width=0.48\linewidth]{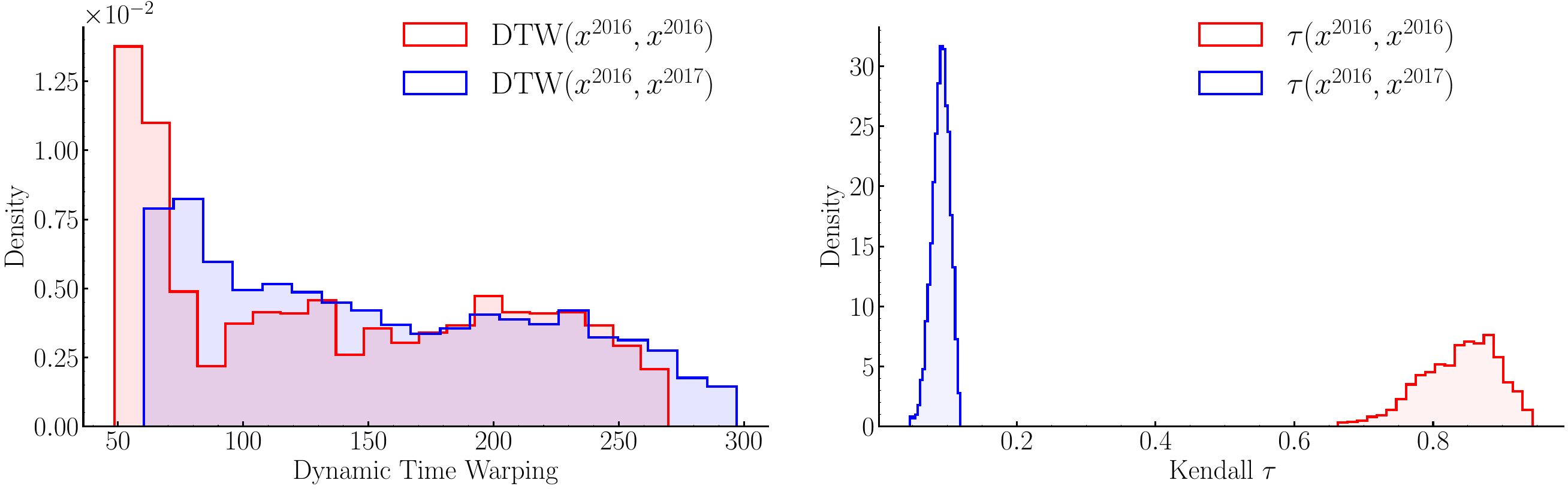}}\\
	\subfigure[Taxi Chicago]{\includegraphics[width=0.48\linewidth]{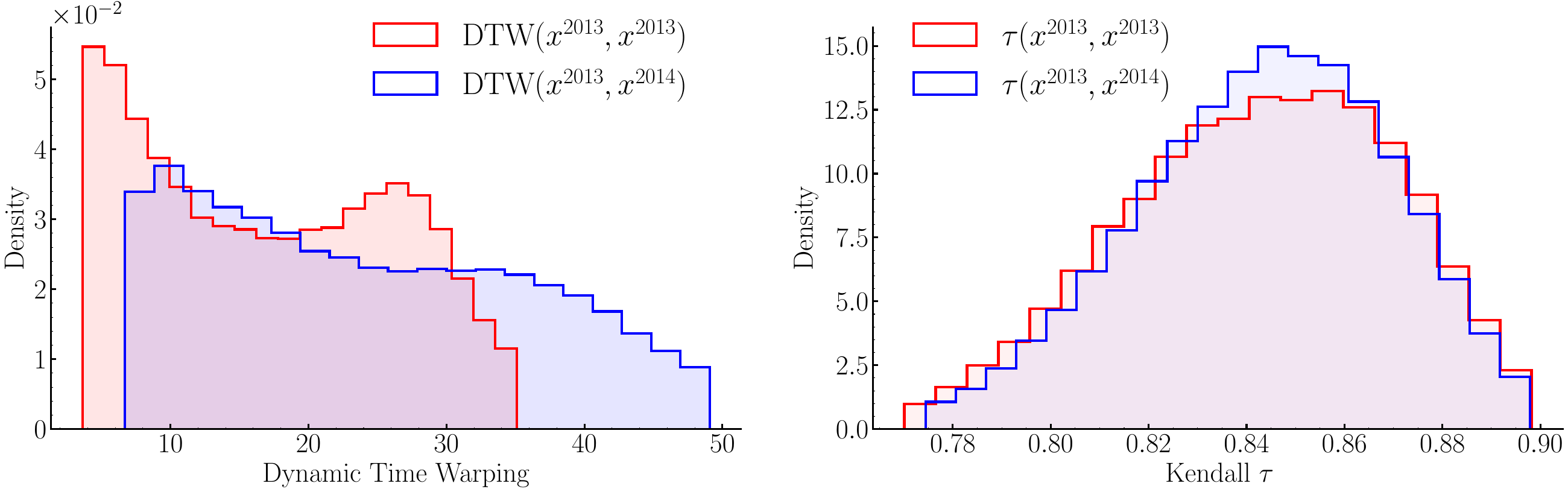}}
	\subfigure[Bike Chicago]{\includegraphics[width=0.48\linewidth]{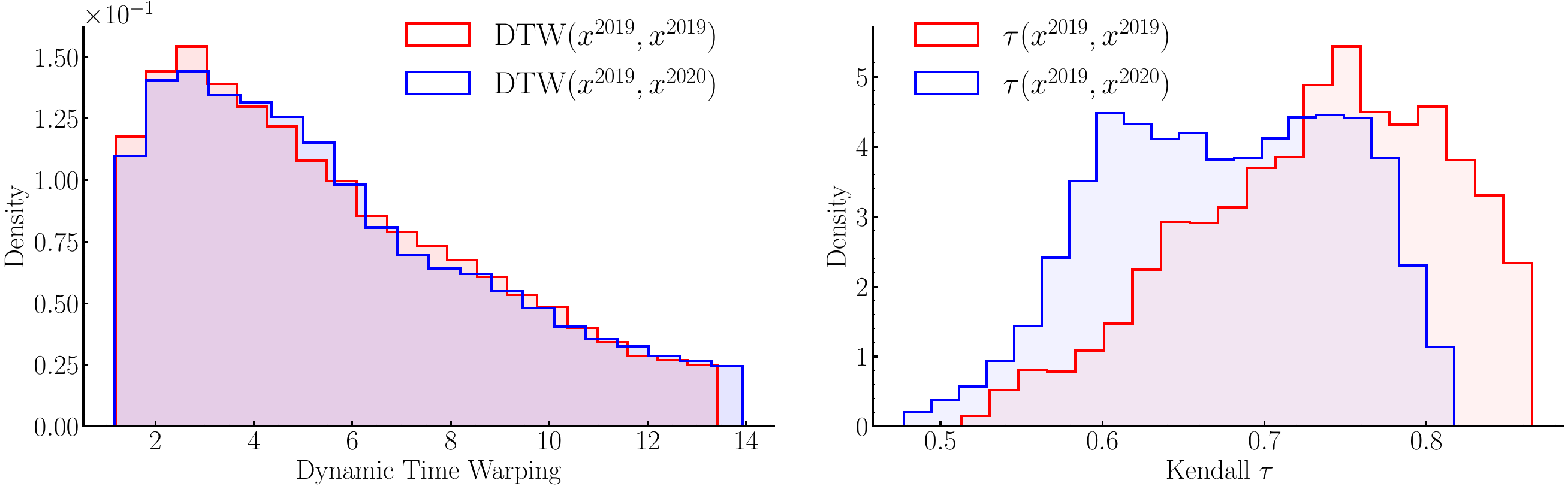}}\\
	\subfigure[Speed NYC]{\includegraphics[width=0.48\linewidth]{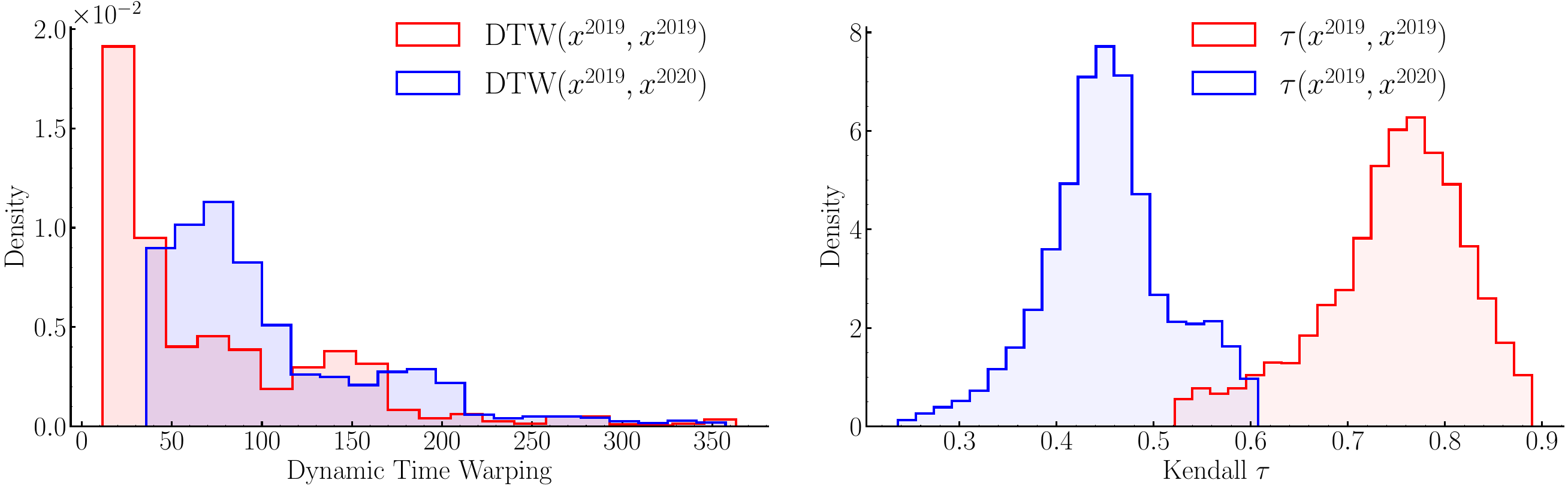}}
	\subfigure[Pedestrians Zurich]{\includegraphics[width=0.48\linewidth]{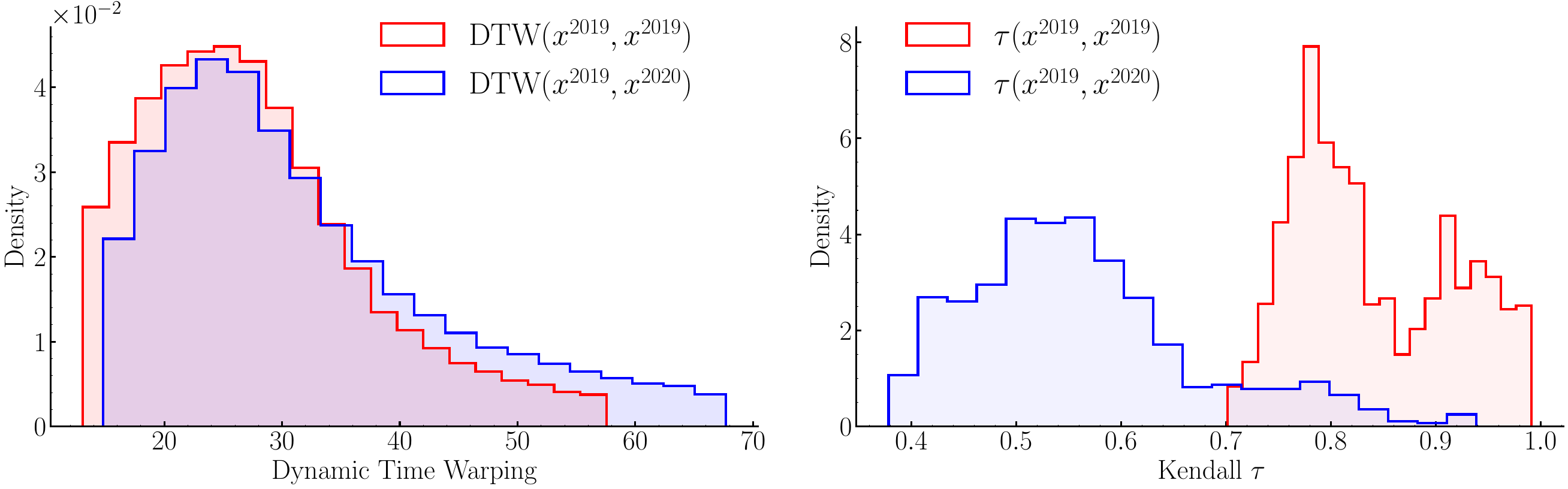}}
	\caption{Compare the DTW  and Kendall's $\tau$ distances for the pedestrian Zurich data  and the 311 service data from New York over the same year and across different years. }
	\label{fig:analysis}
\end{figure*}

\subsection{Data Analysis}\label{sec:2}
In this section, we attempt to explore the differences between in-distribution and out-of-distribution scenarios from a data perspective. To evaluate the similarity of graph relationships, we employ Kendall's $\tau$ coefficient \cite{kendall1938new}, which allows us to quantitatively assess the spatial shifts between in-distribution and out-of-distribution for the same day across different years \cite{edwards2023graphing}. Mathematically, the Kendall's $\tau$ for node $v$ is defined as:
\[
\tau_v =\frac{2}{n(n-1)} \sum_{u \in N(v)} \sum_{i < j} \ \operatorname{sign}\left(x^{(i)}_v - x^{(i)}_u\right) \operatorname{sign}\left(y^{(j)}_v - y^{(j)}_u\right),
\]
where $x_v$ and $y_v$ represents the signal of node $v$ in the same year or across different years.  $N(v) = \{u \in \mathcal{V} \mid (v, u) \in \mathcal{E}\}$ denotes the neighbors of node $v$, $n = |N(v)|$ is the number of neighbors of node $v$, $\mathcal{E}$ represents the edge set of the graph, and the $\operatorname{sign}$ function returns +1 when $x_v > x_u$, -1 when $x_v < x_u$, and 0 when $x_v = x_u$.  Kendall's $\tau$ coefficient ranges from -1 to 1, where a value of 1 indicates perfect agreement between the rankings, -1 indicates perfect disagreement, and 0 suggests no correlation.
We also introduce the Dynamic Time Warping (DTW) \cite{muller2007dynamic} metric to capture the temporal distance between different days.  A DTW value of 0 would mean the sequences are identical, while larger values suggest increasing dissimilarity in the temporal patterns between days. 

In Figure \ref{fig:analysis}, we provide a comprehensive comparison of both intra-year and cross-year results, highlighting the spatiotemporal variations across six urban datasets. To illustrate this, we focus on pedestrian data from Zurich and 311 service reports from New York City. Using in-distribution distances as a baseline, we quantified the extent of spatiotemporal distributional shifts.
Influenced by the COVID-19 pandemic, Zurich’s pedestrian data exhibited relatively stable temporal flow patterns between 2019 and 2020. However, spatial disparities significantly increased, likely due to changes in mobility behaviors, such as the widespread adoption of remote work practices \cite{santana2023covid}. In contrast, 311 service requests in NYC showed minimal spatial variation during the pandemic but experienced significant temporal fluctuations. These temporal changes were primarily driven by the fluctuation in request volumes, with reports indicating a 28\% decrease in 2020 compared to 2019 \cite{Jackson}.
From whole perspectives, it is noteworthy that both the 311 NYC and Taxi Chicago datasets exhibited pronounced temporal OOD trends, whereas Flow SC, Pedestrians Zurich, Bike Chicago, and Speed NYC displayed distinct spatial OOD patterns.
These findings underscore the comprehensiveness of our dataset, which effectively assesses the robustness of existing methods in real-world scenarios. The observed spatiotemporal shifts across different urban datasets highlight the complexity of urban dynamics and emphasize the importance of robust and adaptive modeling approaches in urban computing applications.

\section{Experiments}
\label{sec:5}

\noindent\textbf{Baselines.} ST-OOD benchmark presents an extensive evaluation of traffic forecasting methods, covering a wide range of baselines with publicly accessible implementations. These baselines include traditional, contemporary deep learning, and state-of-the-art models, offering a comprehensive view of advancements in the field.
\begin{itemize}
\item \textbf{LSTM} \cite{hochreiter1997long}: A recurrent neural network architecture, Long Short-Term Memory, is known for effectively capturing and maintaining long-term dependencies in sequential data.

\item \textbf{Graph WaveNet} \cite{wu2019graph}: Stacks Gated Temporal Convolutional Networks (TCN) with Graph Convolutional Networks (GCN) on WaveNet backbone \cite{van2016wavenet} to concurrently capture both spatial and temporal dependencies.

\item \textbf{ASTGCN} \cite{guo2019attention}: The Attention-based Spatial-Temporal Graph Convolutional Network synergizes spatial-temporal attention mechanisms to capture dynamic spatial-temporal patterns in traffic data.

\item \textbf{MTGNN} \cite{wu2020connecting}: Expands Graph WaveNet by incorporating mix-hop propagation layers, dilated inception layers for temporal modeling, and a sophisticated graph learning module.

\item \textbf{STGCN} \cite{yu2018spatio}: A Spatial-Temporal Graph Convolutional Network that models spatial dependencies using graph convolutions and temporal dependencies via 1D convolutions.

\item \textbf{STTN} \cite{xu2020spatial}: The Spatial-Temporal Transformer Network applies the Transformer architecture to simultaneously capture both spatial and temporal dependencies in traffic data.

\item \textbf{STGODE} \cite{fang2021spatial}: Models continuous traffic flow changes over time and space using neural ordinary differential equations.

\item \textbf{DSTAGNN} \cite{lan2022dstagnn}: Integrates dynamic graph learning with attention mechanisms to capture evolving spatial-temporal dependencies within traffic networks.

\item \textbf{D$^2$STGNN} \cite{shao2022decoupled}: Decouples the modeling of spatial and temporal dependencies, addressing the dynamic relationships between traffic sensors.

\item \textbf{STID} \cite{shao2022spatial}: The Spatial-Temporal Identity model combines spatial and temporal embeddings, capturing the unique characteristics of each node and time step for accurate  forecasting.

\item \textbf{MLP}: In this paper, we simplify the STID model by removing the node embedding layer, thus reducing it to a more fundamental architecture: the multilayer perceptron (MLP), a classic feedforward neural network.

\item \textbf{STAEformer} \cite{liu2023spatio}: Enhances standard transformers with adaptive embeddings, efficiently modeling complex spatio-temporal traffic patterns.

\end{itemize}

\noindent\textbf{Implementation Details}. We collected the baseline code from their respective GitHub repositories, performed necessary cleaning, and integrated them into the LargeST\footnote{\url{https://github.com/liuxu77/LargeST/tree/main}} repository to enhance reproducibility and ease of comparison. For model and training configurations, we adhered to the recommended settings provided in their codebases. The experiments were repeated twice using different seeds on a computing server equipped with an Intel(R) Core(TM) i9-10900X CPU @ 3.70GHz, 62 GB RAM, and an NVIDIA RTX 4090 GPU with 48 GB of memory.

\begin{table*}[t!]
	\caption{Performance comparison of ST-OOD prediction in term of MAE, RMSE, and MAPE, respectively.  We also list the average rank for each model. The best model per metric in the six datasets is highlighted in gray. The decline refers to the ratio of performance degradation when comparing OOD performance to in-distribution performance. }
	\label{tbl:res}
	\renewcommand{\arraystretch}{0.9}
	\resizebox{\textwidth}{!}{
		\begin{tabular}{ccccccccccccccccc}
			\toprule
			& \multicolumn{2}{c}{\textbf{311NYC}} & \multicolumn{2}{c}{\textbf{BikeCHI}} & \multicolumn{2}{c}{\textbf{FlowSC}} & \multicolumn{2}{c}{\textbf{PedZurich}} & \multicolumn{2}{c}{\textbf{SpeedNYC}} & \multicolumn{2}{c}{\textbf{TaxiCHI}} & \multicolumn{2}{c}{\textbf{Avg. Rank}} \\
			\cmidrule(lr){2-3} \cmidrule(lr){4-5} \cmidrule(lr){6-7} \cmidrule(lr){8-9} \cmidrule(lr){10-11} \cmidrule(lr){12-13} \cmidrule(lr){14-15} 
			\textbf{Model} & \textbf{IN} & \textbf{OUT} & \textbf{IN} & \textbf{OUT} & \textbf{IN} & \textbf{OUT} & \textbf{IN} & \textbf{OUT} & \textbf{IN} & \textbf{OUT} & \textbf{IN} & \textbf{OUT} & \textbf{IN} & \textbf{OUT}  \\
			\midrule
			STID& 2.09 & 2.61 & 1.30 & 1.91 & 14.27 & 36.76 & 27.85 & 38.54 & 4.65 & 6.19 & 13.73 & 19.74 & 3.67 &\cellcolor{gray!15}   3.67 \\
			MLP& 2.19 & 2.69 & 1.37 & 1.92 & 16.07 & 16.23 & 33.36 & 36.39 & 5.26 & 4.47 & 16.47 & 22.22 & 12.33 & 4.33 \\
			STGODE& 2.10 & 2.61 & 1.25 & 1.94 & 15.45 & 37.39 & 28.54 & 38.84 & 4.82 & 6.62 & 14.13 & 21.71 & 6.17 & 5.67 \\
			WaveNet& 2.18 & 2.70 & 1.37 & 1.94 & 16.39 & 16.73 & 31.41 & 38.34 & 5.19 & 4.47 & 15.72 & 23.98 & 11.17 & 5.67 \\
			LSTM& 2.18 & 2.70 & 1.36 & 1.92 & 16.47 & 16.84 & 31.16 & 37.43 & 5.19 & 4.50 & 16.71 & 26.23 & 11.33 & 6.00 \\
			GWNet&  2.06 &  2.60 & 1.29 & 2.04 & 14.66 & 25.41 & 29.98 & 48.53 & 4.77 & 6.79 & 13.74 & 20.97 & 4.33 & 6.17 \\
			D$^2$STGNN& 2.08 & 2.64 & 1.22 & 2.07 & 14.74 & 38.70 & 27.82 & 46.70 & 4.56 & 6.39 & 13.81 & 20.74 & \cellcolor{gray!15}  2.50 & 6.83 \\
			STAEformer& 2.09 & 2.64 & 1.32 & 2.14 & 13.52 & 31.39 & 28.65 & 41.85 & 4.70 & 6.15 & 13.83 & 21.78 & 4.83 & 7.33 \\
			DAAGCN& 2.09 & 2.61 & 1.26 & 2.13 & 14.73 & 38.92 & 30.92 & 46.89 & 4.72 & 6.43 & 14.42 & 22.26 & 5.67 & 7.83 \\
			MTGNN& 2.08 & 2.61 & 1.23 & 2.13 & 15.12 & 45.97 & 29.92 & 48.52 & 4.72 & 6.54 & 13.80 & 20.98 & 3.83 & 7.83 \\
			STTN& 2.10 & 2.63 & 1.25 & 1.98 & 15.19 & 36.80 & 32.32 & 52.90 & 4.93 & 6.69 & 14.11 & 22.02 & 7.67 & 8.17 \\
			STGCN& 2.14 & 2.61 & 1.30 & 2.08 & 15.76 & 69.17 & 32.06 & 54.53 & 4.68 & 7.46 & 14.58 & 22.99 & 8.33 & 10.67 \\
			AGCRN& 2.11 & 2.62 & 1.27 & 2.08 & 15.23 & 48.53 & 33.67 & 57.94 & 4.89 & 7.01 & 15.35 & 24.93 & 9.17 & 10.83 \\
			\midrule
			\multicolumn{1}{c}{\textbf{\textcolor{red}{Decline  (\%)}}} & \multicolumn{2}{c}{\bf \textcolor{red}{24.75\% $\bm{\downarrow}$} } & \multicolumn{2}{c}{\bf \textcolor{red}{56.73\% $\bm{\downarrow}$} } & \multicolumn{2}{c}{\bf \textcolor{red}{133.92\% $\bm{\downarrow}$} }  & \multicolumn{2}{c}{\bf \textcolor{red}{47.78\% $\bm{\downarrow}$} }  &\multicolumn{2}{c}{\bf \textcolor{red}{27.26\% $\bm{\downarrow}$} }   &\multicolumn{2}{c}{\bf \textcolor{red}{52.68\% $\bm{\downarrow}$} }  & \multicolumn{2}{c}{-} \\
			\bottomrule
		\end{tabular}
	}
	
	\resizebox{\textwidth}{!}{
		\begin{tabular}{ccccccccccccccccc}
			\toprule
			& \multicolumn{2}{c}{\textbf{311NYC}} & \multicolumn{2}{c}{\textbf{BikeCHI}} & \multicolumn{2}{c}{\textbf{FlowSC}} & \multicolumn{2}{c}{\textbf{PedZurich}} & \multicolumn{2}{c}{\textbf{SpeedNYC}} & \multicolumn{2}{c}{\textbf{TaxiCHI}} & \multicolumn{2}{c}{\textbf{Avg. Rank}} \\
			\cmidrule(lr){2-3} \cmidrule(lr){4-5} \cmidrule(lr){6-7} \cmidrule(lr){8-9} \cmidrule(lr){10-11} \cmidrule(lr){12-13} \cmidrule(lr){14-15} 
			\textbf{Model} & \textbf{IN} & \textbf{OUT} & \textbf{IN} & \textbf{OUT} & \textbf{IN} & \textbf{OUT} & \textbf{IN} & \textbf{OUT} & \textbf{IN} & \textbf{OUT} & \textbf{IN} & \textbf{OUT} & \textbf{IN} & \textbf{OUT}  \\
			\midrule
			STID& 3.15 & 4.45 & 2.42 & 3.39 & 23.21 & 54.35 & 51.20 & 61.02 & 7.58 & 9.43 & 45.15 & 71.81 & \cellcolor{gray!15}  2.00 & \cellcolor{gray!15}  3.00 \\
			MLP& 3.29 & 4.51 & 2.73 & 3.53 & 26.28 & 27.81 & 60.40 & 63.10 & 8.28 & 7.66 & 52.32 & 77.47 & 11.67 & 4.00 \\
			STGODE& 3.16 & 4.44 & 2.37 & 3.46 & 24.37 & 55.92 & 52.70 & 62.18 & 7.66 & 9.79 & 46.76 & 80.31 & 4.67 & 4.17 \\
			LSTM& 3.31 & 4.55 & 2.73 & 3.52 & 26.52 & 28.44 & 56.54 & 64.42 & 8.29 & 7.79 & 55.40 & 94.51 & 12.00 & 6.33 \\
			WaveNet& 3.30 & 4.53 & 2.73 & 3.56 & 26.53 & 28.56 & 57.22 & 66.12 & 8.21 & 7.67 & 50.87 & 85.12 & 11.33 & 6.50 \\
			GWNet& 3.16 & 4.47 & 2.51 & 3.99 & 23.49 & 38.16 & 54.90 & 78.10 & 7.66 & 9.95 & 45.61 & 76.51 & 4.50 & 6.67 \\
			MTGNN& 3.15 & 4.45 & 2.35 & 4.13 & 24.29 & 66.08 & 55.19 & 78.37 & 7.70 & 10.01 & 45.18 & 76.52 & 4.17 & 7.67 \\
			D$^2$STGNN& 3.17 & 4.51 & 2.37 & 4.14 & 23.75 & 58.80 & 52.56 & 76.01 & 7.54 & 9.89 & 45.00 & 74.13 & 3.00 & 7.67 \\
			STAEformer& 3.18 & 4.51 & 2.64 & 4.48 & 23.62 & 47.45 & 52.62 & 65.83 & 7.68 & 9.46 & 46.64 & 80.78 & 5.83 & 7.67 \\
			DAAGCN& 3.17 & 4.45 & 2.45 & 4.14 & 24.30 & 56.93 & 55.64 & 70.58 & 7.71 & 9.91 & 50.32 & 84.47 & 7.17 & 8.00 \\
			STTN& 3.17 & 4.47 & 2.38 & 3.69 & 24.10 & 54.42 & 56.65 & 80.36 & 8.05 & 10.04 & 48.25 & 83.27 & 6.83 & 8.33 \\
			STGCN& 3.21 & 4.45 & 2.52 & 4.15 & 24.91 & 100.52 & 58.29 & 85.82 & 7.63 & 10.66 & 48.64 & 83.17 & 8.50 & 10.33 \\
			AGCRN& 3.18 & 4.47 & 2.44 & 3.99 & 24.51 & 69.27 & 58.52 & 87.13 & 7.90 & 10.16 & 51.95 & 90.20 & 9.33 & 10.67 \\
			\midrule
			\multicolumn{1}{c}{\textbf{\textcolor{red}{Decline  (\%)}}} & \multicolumn{2}{c}{\bf \textcolor{red}{40.06\% $\bm{\downarrow}$} } & \multicolumn{2}{c}{\bf \textcolor{red}{54.25\% $\bm{\downarrow}$} } & \multicolumn{2}{c}{\bf \textcolor{red}{116.44\% $\bm{\downarrow}$} }  & \multicolumn{2}{c}{\bf \textcolor{red}{30.00\% $\bm{\downarrow}$} }  &\multicolumn{2}{c}{\bf \textcolor{red}{20.60\% $\bm{\downarrow}$} }   &\multicolumn{2}{c}{\bf \textcolor{red}{67.45\% $\bm{\downarrow}$} }  & \multicolumn{2}{c}{-} \\
			\bottomrule
		\end{tabular}
	}	
	\resizebox{\textwidth}{!}{
		\begin{tabular}{ccccccccccccccccc}
			\toprule
			& \multicolumn{2}{c}{\textbf{311NYC}} & \multicolumn{2}{c}{\textbf{BikeCHI}} & \multicolumn{2}{c}{\textbf{FlowSC}} & \multicolumn{2}{c}{\textbf{PedZurich}} & \multicolumn{2}{c}{\textbf{SpeedNYC}} & \multicolumn{2}{c}{\textbf{TaxiCHI}} & \multicolumn{2}{c}{\textbf{Avg. Rank}} \\
			\cmidrule(lr){2-3} \cmidrule(lr){4-5} \cmidrule(lr){6-7} \cmidrule(lr){8-9} \cmidrule(lr){10-11} \cmidrule(lr){12-13} \cmidrule(lr){14-15} 
			\textbf{Model} & \textbf{IN} & \textbf{OUT} & \textbf{IN} & \textbf{OUT} & \textbf{IN} & \textbf{OUT} & \textbf{IN} & \textbf{OUT} & \textbf{IN} & \textbf{OUT} & \textbf{IN} & \textbf{OUT} & \textbf{IN} & \textbf{OUT}  \\
			\midrule
		D2STGNN& 57.41\% & 54.87\% & 44.88\% & 56.80\% & 10.58\% & 38.75\% & 52.99\% & 105.76\% & 23.76\% & 25.42\% & 35.74\% & 39.96\% & \cellcolor{gray!15} 3.83 &\cellcolor{gray!15} 5.17 \\
		STID& 60.14\% & 56.36\% & 53.31\% & 56.85\% & 9.34\% & 25.49\% & 64.38\% & 115.60\% & 25.87\% & 24.40\% & 38.74\% & 40.06\% & 6.33 &\cellcolor{gray!15} 5.17 \\
		MLP& 63.27\% & 61.89\% & 51.02\% & 55.79\% & 10.33\% & 9.24\% & 72.83\% & 82.55\% & 27.21\% & 18.34\% & 51.75\% & 53.72\% & 11.50 & 5.33 \\
		LSTM& 60.48\% & 59.50\% & 47.94\% & 54.01\% & 10.57\% & 9.56\% & 65.80\% & 82.45\% & 26.45\% & 18.50\% & 49.37\% & 54.05\% & 8.83 & 5.33 \\
		WAVENET& 61.60\% & 60.84\% & 49.55\% & 54.92\% & 10.56\% & 9.50\% & 66.46\% & 84.69\% & 26.74\% & 18.40\% & 43.97\% & 47.93\% & 10.00 & 5.33 \\
		STAEFORMER& 57.58\% & 55.37\% & 50.70\% & 61.87\% & 8.86\% & 28.65\% & 65.60\% & 127.35\% & 24.96\% & 24.84\% & 33.81\% & 36.63\% & 4.00 & 6.00 \\
		GWNET& 55.56\% & 53.95\% & 49.86\% & 61.25\% & 9.78\% & 21.08\% & 63.04\% & 128.80\% & 25.54\% & 26.11\% & 40.11\% & 43.56\% & 4.50 & 6.50 \\
		STGODE& 60.40\% & 57.20\% & 48.40\% & 58.02\% & 10.13\% & 34.44\% & 65.54\% & 117.00\% & 26.38\% & 24.65\% & 35.78\% & 37.68\% & 6.67 & 6.50 \\
		STTN& 59.07\% & 56.25\% & 48.27\% & 57.41\% & 9.84\% & 25.42\% & 96.46\% & 194.48\% & 27.45\% & 27.30\% & 35.23\% & 45.88\% & 6.83 & 7.83 \\
		MTGNN& 58.45\% & 56.69\% & 45.78\% & 59.77\% & 10.05\% & 49.68\% & 65.76\% & 130.81\% & 25.93\% & 27.34\% & 35.09\% & 37.56\% & 4.33 & 8.83 \\
		STGCN& 61.36\% & 56.60\% & 50.16\% & 58.99\% & 10.30\% & 81.22\% & 96.83\% & 234.73\% & 24.20\% & 27.19\% & 35.33\% & 38.01\% & 8.00 & 9.17 \\
		DAAGCN& 58.55\% & 56.51\% & 48.43\% & 63.17\% & 10.04\% & 43.92\% & 79.84\% & 156.02\% & 26.26\% & 26.41\% & 43.38\% & 44.63\% & 7.33 & 9.33 \\
		AGCRN& 59.98\% & 57.12\% & 48.39\% & 59.10\% & 10.16\% & 51.36\% & 108.95\% & 254.95\% & 26.89\% & 26.49\% & 41.72\% & 47.32\% & 8.83 & 10.50 \\
			\midrule
			\multicolumn{1}{c}{\textbf{\textcolor{red}{Decline  (\%)}}} & \multicolumn{2}{c}{\bf \textcolor{red}{-3.97\% $\bm{\downarrow}$} } & \multicolumn{2}{c}{\bf \textcolor{red}{19.27\% $\bm{\downarrow}$} } & \multicolumn{2}{c}{\bf \textcolor{red}{228.62\% $\bm{\downarrow}$} }  & \multicolumn{2}{c}{\bf \textcolor{red}{84.20\% $\bm{\downarrow}$} }  &\multicolumn{2}{c}{\bf \textcolor{red}{-6.23\% $\bm{\downarrow}$} }   &\multicolumn{2}{c}{\bf \textcolor{red}{9.30\% $\bm{\downarrow}$} }  & \multicolumn{2}{c}{-} \\
			\bottomrule
		\end{tabular}
	}
\end{table*}

\begin{table*}[t!]
	\caption{Performance comparison with  spatiotemporal OOD model. Best performance per benchmark highlighted in gray.  }
	\label{tbl:ood}
	\renewcommand{\arraystretch}{1.}
	\resizebox{\textwidth}{!}{
		\begin{tabular}{ccccccccccccc}
			\toprule
			& \multicolumn{2}{c}{\textbf{311NYC}} & \multicolumn{2}{c}{\textbf{BikeCHI}} & \multicolumn{2}{c}{\textbf{FlowSC}} & \multicolumn{2}{c}{\textbf{PedZurich}} & \multicolumn{2}{c}{\textbf{SpeedNYC}} & \multicolumn{2}{c}{\textbf{TaxiCHI}} \\
			\cmidrule(lr){2-3} \cmidrule(lr){4-5} \cmidrule(lr){6-7} \cmidrule(lr){8-9} \cmidrule(lr){10-11} \cmidrule(lr){12-13} 
			\textbf{Model} & \textbf{IN} & \textbf{OUT} & \textbf{IN} & \textbf{OUT} & \textbf{IN} & \textbf{OUT} & \textbf{IN} & \textbf{OUT} & \textbf{IN} & \textbf{OUT} & \textbf{IN} & \textbf{OUT}   \\
			\midrule
			STID& \cellcolor{gray!15} 2.09 &  2.61 & \cellcolor{gray!15} 1.30 & \cellcolor{gray!15} 1.91 & \cellcolor{gray!15} 14.27 & 36.76 & \cellcolor{gray!15} 27.85 &  38.54 & \cellcolor{gray!15} 4.65 &  6.19 & \cellcolor{gray!15} 13.73 & 19.74  \\
			w.o. Node & 2.19 & 2.69 & 1.37 & 1.92 & 16.07 &\cellcolor{gray!15} 16.23 & 33.36 & \cellcolor{gray!15} 36.39 & 5.26 &\cellcolor{gray!15} 4.47 & 16.47 & 22.22 \\
			\hline
			CaST & 2.86 & 2.78 & 1.78 & 3.87 & 23.73 & 22.52 & 40.91 & 62.15 & 6.93 & 7.13 & 23.11 & 22.24 \\
			CauSTG & 3.14 & 3.05 & 1.95 & 4.25 & 26.04 & 24.72 & 44.89 & 68.23 & 7.61 & 7.83 & 25.36 & 24.41 \\
			STONE & 2.15 & \cellcolor{gray!15} 2.20 & 1.34 & 3.06 & 17.84 &  17.85 & 30.76 & 49.22  & 5.21 &  5.65 & 17.37  & \cellcolor{gray!15} 17.62  \\
			\bottomrule
		\end{tabular}
	}
	
	\resizebox{\textwidth}{!}{
		\begin{tabular}{ccccccccccccccc}
			\toprule
			& \multicolumn{2}{c}{\textbf{311NYC}} & \multicolumn{2}{c}{\textbf{BikeCHI}} & \multicolumn{2}{c}{\textbf{FlowSC}} & \multicolumn{2}{c}{\textbf{PedZurich}} & \multicolumn{2}{c}{\textbf{SpeedNYC}} & \multicolumn{2}{c}{\textbf{TaxiCHI}}  \\
			\cmidrule(lr){2-3} \cmidrule(lr){4-5} \cmidrule(lr){6-7} \cmidrule(lr){8-9} \cmidrule(lr){10-11} \cmidrule(lr){12-13} 
			\textbf{Model} & \textbf{IN} & \textbf{OUT} & \textbf{IN} & \textbf{OUT} & \textbf{IN} & \textbf{OUT} & \textbf{IN} & \textbf{OUT} & \textbf{IN} & \textbf{OUT} & \textbf{IN} & \textbf{OUT}  \\
			\midrule
			STID& \cellcolor{gray!15} 3.15 & 4.45 & \cellcolor{gray!15} 2.42 & \cellcolor{gray!15} 3.39 & \cellcolor{gray!15} 23.21 & 54.35 & \cellcolor{gray!15} 51.20 & \cellcolor{gray!15} 61.02 & \cellcolor{gray!15} 7.58 &  9.43 & \cellcolor{gray!15} 45.15 & 71.81  \\
			w.o. Node& 3.29 & 4.51 & 2.73 & 3.53 & 26.28 &\cellcolor{gray!15} 27.81 & 60.40 & 63.10 & 8.28 &\cellcolor{gray!15} 7.66 & 52.32 & 77.47  \\
			\hline
			CaST & 4.26 & 4.19 & 3.41 & 6.92 & 36.95 & 36.32 & 73.63 & 105.03 & 11.00 & 10.71 & 69.03 & 66.37 \\
			CauSTG & 4.67 & 4.60 & 3.74 & 7.60 & 40.56 & 39.86 & 80.83 & 115.32 & 12.08 & 11.76 & 75.79 & 72.87 \\
			STONE & 3.20 &  \cellcolor{gray!15} 3.32 & 2.56 & 5.48 & 27.77 &  28.78 & 55.35 & 83.22  & 8.27 &8.49 &  51.90 & \cellcolor{gray!15} 52.59 \\ 
			\hline
		\end{tabular}
	}	
	\resizebox{\textwidth}{!}{
		\begin{tabular}{ccccccccccccccc}
			\toprule
			& \multicolumn{2}{c}{\textbf{311NYC}} & \multicolumn{2}{c}{\textbf{BikeCHI}} & \multicolumn{2}{c}{\textbf{FlowSC}} & \multicolumn{2}{c}{\textbf{PedZurich}} & \multicolumn{2}{c}{\textbf{SpeedNYC}} & \multicolumn{2}{c}{\textbf{TaxiCHI}} \\
			\cmidrule(lr){2-3} \cmidrule(lr){4-5} \cmidrule(lr){6-7} \cmidrule(lr){8-9} \cmidrule(lr){10-11} \cmidrule(lr){12-13}
			\textbf{Model} & \textbf{IN} & \textbf{OUT} & \textbf{IN} & \textbf{OUT} & \textbf{IN} & \textbf{OUT} & \textbf{IN} & \textbf{OUT} & \textbf{IN} & \textbf{OUT} & \textbf{IN} & \textbf{OUT}  \\
			\midrule
			STID & \cellcolor{gray!15} 60.14\% & 56.36\% & 53.31\% & 56.85\% &\cellcolor{gray!15}  9.34\% & 25.49\% & \cellcolor{gray!15} 64.38\% & 115.60\% &\cellcolor{gray!15} 25.87\% & 24.40\% &\cellcolor{gray!15} 38.74\% &\cellcolor{gray!15} 40.06\%  \\
			w.o. Node& 63.27\% & 61.89\% & \cellcolor{gray!15} 51.02\% &\cellcolor{gray!15} 55.79\% & 10.33\% & \cellcolor{gray!15} 9.24\% & 72.83\% & 82.55\% & 27.21\% &\cellcolor{gray!15} 18.34\% & 51.75\% & 53.72\%  \\
			\hline
		CaST & 86.24\% & 64.73\% & 68.91\% & 73.56\% & 23.18\% & 43.05\% & 88.39\% & 96.82\% & 35.47\% & 114.69\% & 74.31\% & 98.95\% \\
		CauSTG & 95.63\% & 71.09\% & 75.42\% & 80.17\% & 25.84\% & 47.36\% & 96.51\% & 105.28\% & 38.75\% & 125.93\% & 82.04\% & 108.70\% \\
		STONE & 65.12\% &\cellcolor{gray!15}  51.79\% &  51.38\% & 58.96\% & 17.25\% & 34.61\% & 66.87\% & \cellcolor{gray!15} 76.43\%  & 27.59\% & 90.21\% & 56.34\% & 78.08\% \\
			\bottomrule
		\end{tabular}
	}
\end{table*}

\noindent\textbf{Evaluation Metrics}. We used three commonly employed metrics in prediction tasks to evaluate model performance: Mean Absolute Error (MAE), Root Mean Squared Error (RMSE), and Mean Absolute Percentage Error (MAPE).  Note that the maximum batch size was set to 64. If a model failed to run under this setting, we gradually reduced the batch size until it fully utilized the memory on the 4090 GPU.

\noindent\textbf{Comparison of Baseline Performance on In Distribution Scenario.} Table \ref{tbl:res} lists the average MAE, RMSE, and MAPE values for 12-step predictions across all forecast horizons. First, focusing on in-distribution results, we observe that TCN-based methods (e.g., GWNet and MTGNN) consistently outperform their predecessors, STGCN and AGCRN, across different datasets.  A common characteristic of these methods is their use of real world topology graph and adaptive adjacency matrices, enabling the model to capture a global receptive field. Notably, based on the MAE and MAPE metric, D$^2$STGNN demonstrates outstanding performance across all datasets, indicating that the combination of dynamic spatial graphs and adaptive static graphs is generally more effective than standalone spatial attention mechanisms for small scale values. We hypothesize that this is because the integration of dynamic and static graphs allows the model to learn richer spatial relationships, preventing overfitting. Based on the RMSE metric, STID exhibits strong performance, which we attribute to its use of the entire time series during encoding, better capturing trends and other factors for long-term forecasting.

\noindent\textbf{Comparison of Baseline Performance on OOD Scenario.} Considering the out-of-distribution results, we can observe that STID demonstrates strong performance, achieving the highest OOD prediction accuracy on most datasets, with an average MAE ranking of 3.67 and an average RMSE ranking of 3.00. This highlights the effectiveness and potential of the simple MLP architecture in spatiotemporal forecasting tasks. Additionally, early models that integrate real-world topological graphs and semantic graphs with ordinary differential equations (e.g., STGODE) remain competitive in OOD scenario, with an average MAE ranking of 5.67. Furthermore, methods based on WaveNet \cite{oord2016WaveNET} continue to exhibit robust performance in modeling spatiotemporal dynamics. A noteworthy finding is that all models experience a significant performance drop when transitioning from IN to OUT predictions. The RMSE performance degradation ranges from 40.06\% (SpeedNYC) to 116.44\% (FlowSC), reflecting the inherent challenge of maintaining generalization in spatiotemporal forecasting across different distributions.

\noindent\textbf{Comparison of spatiotemporal OOD Methods Performance.} Based on \tableref{tbl:res}, we selected the best-performing model, STID, and compared it with current leading spatiotemporal OOD models. 
The results are shown in \tableref{tbl:ood}, where w.o. Adp indicates the exclusion of node embedding, meaning no inductive biases are introduced.
We found that removing node embedding significantly improved STID's performance on the FlowSC and SpeedNYC datasets. Furthermore, STID outperformed current OOD-focused methods across most tasks.
It is evident that mainstream methods (e.g., STONE) mainly achieve better generalization by underfitting the spatiotemporal model. Specifically, from datasets like 311NYC, FlowSC, and SpeedNYC, we observe that STONE performs significantly worse than STID on in-distribution data, allowing it to maintain similar performance in out-of-distribution settings. However, STONE exhibits instability in capturing spatiotemporal relationships, as shown in datasets like BikeCHI and PedZurich, where its performance is inferior to STID on both in-distribution and out-of-distribution data. This raises concerns about the effectiveness of current spatiotemporal OOD methods. We argue that a robust OOD method must meet two criteria: strong in-distribution performance and stable out-of-distribution performance. However, due to the lack of proper benchmarks, current spatiotemporal methods are either validated over short time spans or rely on manual node adjustments. As shown in the table, such evaluation approaches are incomplete and insufficient.

\begin{figure*}[t]
	\centering
	\includegraphics[width=1\linewidth]{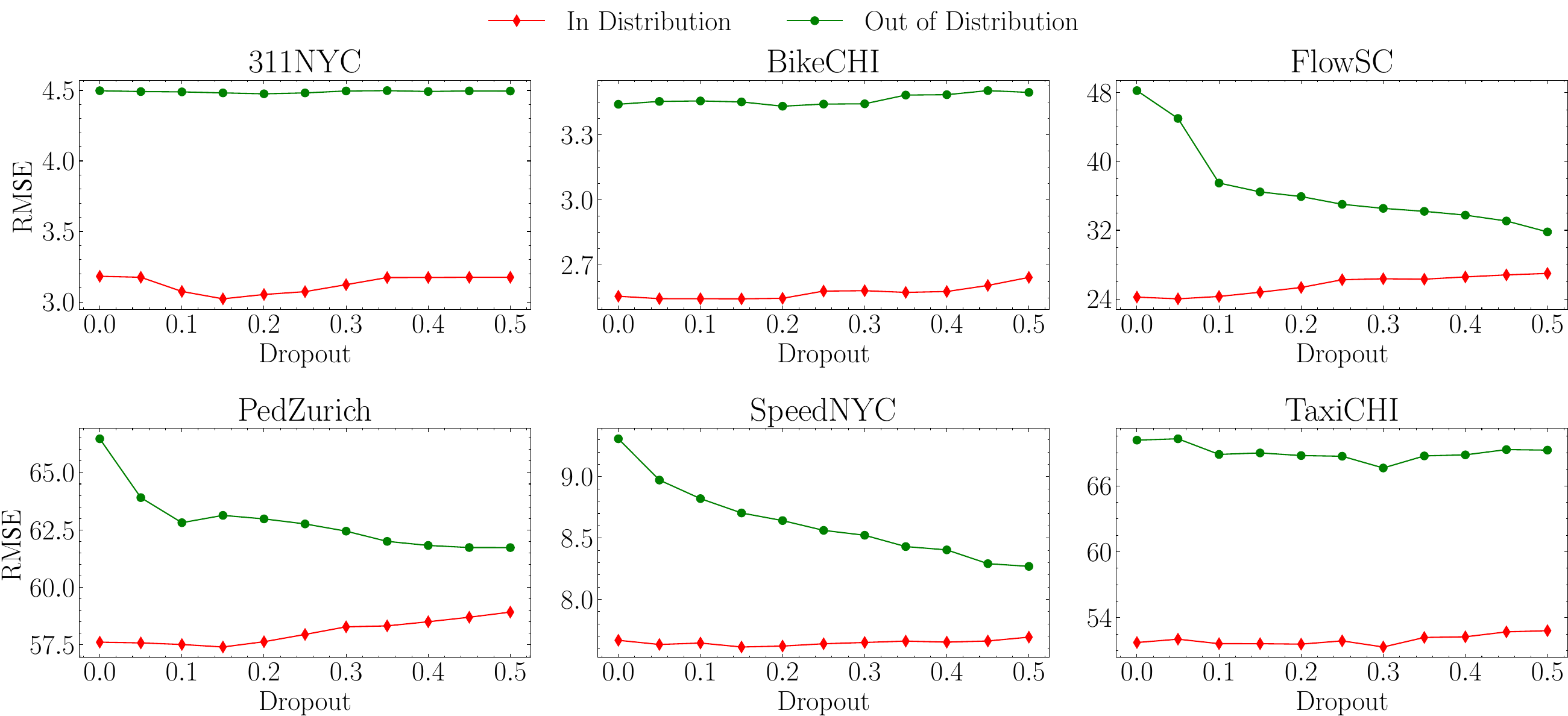}
	\caption{We applied varying proportions of dropout to the node embedding in STID to mitigate the effect of inductive biases, observing its impact on both in and out-of-distribution performance.}
	\label{fig:drop_stid}
\end{figure*}

\noindent\textbf{Investigating the Effect of Dropout on STID.} Based on the observations from \tableref{tbl:ood}, we found that removing node embeddings improved model performance on certain datasets. Specifically, from \figref{fig:analysis}, we observed that these improvements occurred primarily in scenarios where spatial information drift was dominant, such as in FlowSC, PedZurich and SpeedNYC. This suggests that the model may have been overly reliant on node embeddings. To further investigate, we conducted experiments with dropout rates ranging from 0 to 0.5 (in increments of 0.05) across six datasets. In \figref{fig:drop_stid}, we found that moderate dropout significantly enhanced the model’s generalization, though it slightly reduced its in-distribution performance. For instance, on FlowSC, applying moderate dropout reduced the out-of-distribution RMSE error from 54 to 24, while the in-distribution performance improved by only 3 points. Similarly, on SpeedNYC, dropout had almost no impact on in-distribution results but reduced the RMSE by 1 point in out-of-distribution settings. These findings indicate that the model tends to over-rely on the inductive bias introduced by node embeddings, and that applying moderate dropout can improve the model's generalization capabilities. By using dropout, we demonstrate that mitigating the effects of inductive biases can significantly enhance model robustness. However, further research is needed to achieve a balance between in-distribution and out-of-distribution performance.

\begin{figure*}[t]
	\centering
	\includegraphics[width=1\linewidth]{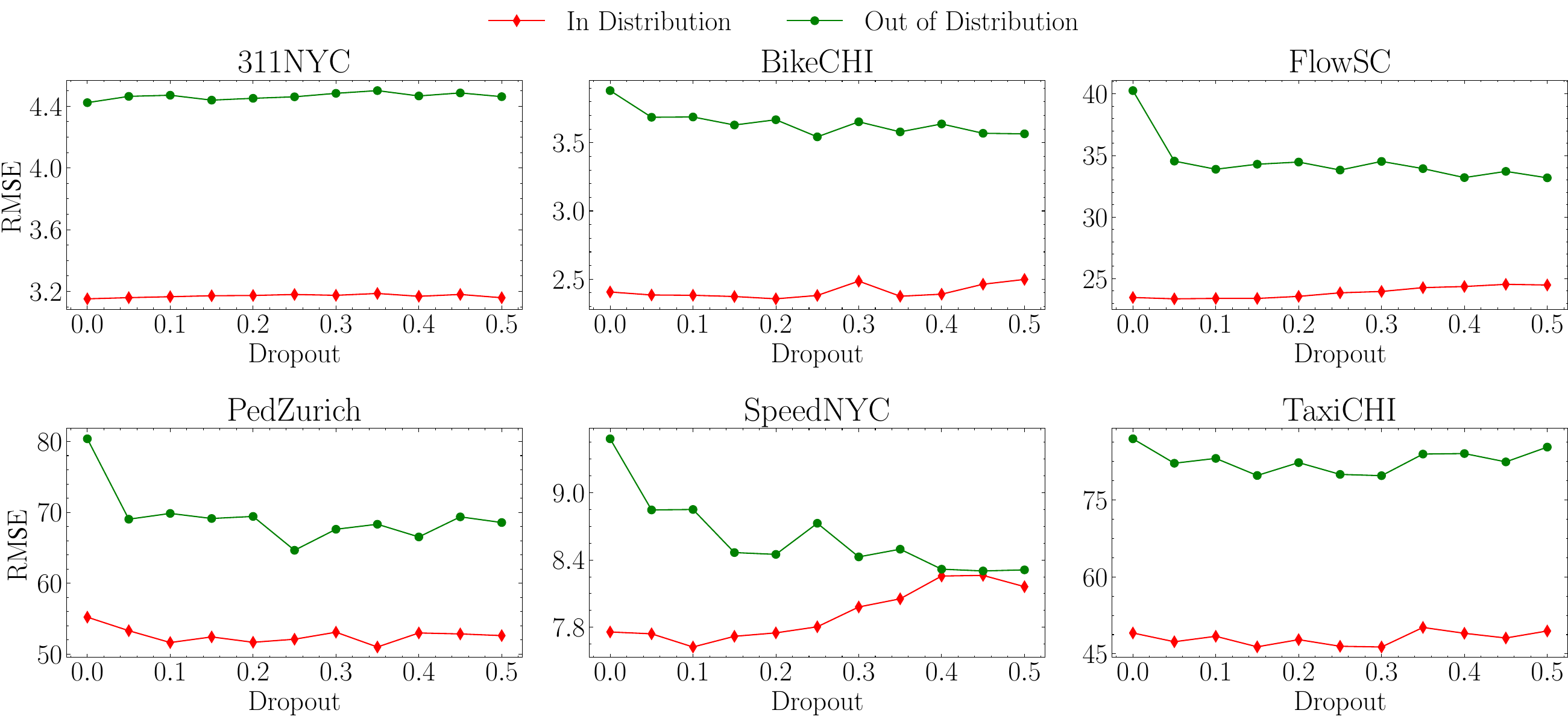}
	\caption{We applied varying proportions of dropout to the node embedding in GWNet to mitigate the effect of inductive biases, observing its impact on both in and out-of-distribution performance.}
	\label{fig:drop_gwnet}
\end{figure*}

\noindent\textbf{Investigating the Effect of Dropout on GWNet.} Beyond validating the efficacy of dropout in mitigating inductive biases on STID, we further conducted experiments on GWNet and illustrated the result in \figref{fig:drop_gwnet}. Using the same six datasets, we varied the dropout rate from 0 to 0.5 in increments of 0.05. Our findings revealed that moderate dropout significantly enhanced GWNet's generalization ability, albeit at the cost of a slight decrease in in-distribution performance. However, in contrast to STID, on FlowSC, GWNet achieved a substantial improvement in out-of-distribution  performance with minimal impact on in-distribution results, attesting to the robustness of the WaveNet architecture. Similarly, on SpeedNYC, a small amount of dropout led to a significant boost in the model's generalization ability. The results from both STID and GWNet suggest that exploring more robust model architectures is a promising avenue for future research.

\section{Discussion}
As illustrated in \tableref{tbl:res}, MLPs, GNNs and Transformer display fundamentally different inductive biases, which significantly influence their generalization in spatiotemporal forecasting. STID employs MLPs based on the assumption of independence among input features, with a bias toward learning global relationships while disregarding the inherent structure of the data. STID relies exclusively on node embeddings to distinguish nodes, thereby reducing the introduction of spatial inductive biases. In contrast, GNNs and Transformer explicitly exploit the relational structure of ST data, learning node representations through local \cite{kipf2016semi} or global neighborhood aggregation \cite{vaswani2017attention}. This structure-aware bias makes GNNs and Transformer well-suited to capturing complex dependencies in the data, though it may also lead to an over-reliance on neighborhood, exacerbating issues related to spurious correlations \cite{srinivasan2019equivalence}. For example, in recommender systems, GNNs may reinforce social biases embedded in interaction graphs, leading to biased recommendations \cite{rahman2019fairwalk}. Moreover, when handling dynamic graphs, GNNs may struggle to distinguish between transient connections and persistent relationships, potentially amplifying short-term spurious correlations \cite{xu2019spatio}. Understanding these inductive biases is essential for selecting appropriate models and designing learning strategies for spatiotemporal tasks, especially in the presence of complex data prone to spurious correlations.

Furthermore, our findings suggest that the lack of a rigorous benchmark for spatiotemporal generalization has hindered the proper evaluation of models in OOD scenarios. Existing models either rely on limited datasets \cite{xia2024deciphering}, introduce artificial noise \cite{zhou2023maintaining}, or remove nodes selectively \cite{wang2024stone}, resulting in inadequate testing methodologies. Using the ST-OOD dataset, we have uncovered significant limitations in current spatiotemporal OOD models, particularly their suboptimal in-distribution performance and instability in OOD scenarios, sometimes yielding worse results than existing methods. Additionally, there is no unified strategy that can be applied seamlessly across current models. Given that many approaches have been validated on conventional benchmarks, addressing these shortcomings is critical for developing more robust spatiotemporal models. Moreover, we challenge the assumption that invariant environmental features can be consistently learned, as spatiotemporal invariance is inherently difficult to define. For instance, predicting future traffic disruptions caused by urban developments or new infrastructure is highly uncertain. We hypothesize that the failure of current models is largely due to this unpredictability, and argue that enhancing models’ extrapolation capabilities is more crucial than pursuing spatiotemporal invariances \cite{krueger2021out}.

\section{Future Opportunities and Limitations}

To advance research on OOD spatiotemporal prediction, we introduce ST-OOD as a new benchmark dataset. It comprises six types of urban data: shared bikes, 311 services, pedestrian counts, traffic speed, traffic flow, and ride-hailing demand. Based on thorough data analysis and extensive experimental results, we highlight the following research opportunities for future exploration:  
\begin{itemize}[leftmargin=6.5mm]
	\item[\ding{117}] \textbf{Integrating  graph and time-series  OOD theory into spatiotemporal prediction. }Recent advances in out-of-distribution generalization on graphs \cite{li2022out}, which extends beyond the in-distribution assumption, have gained significant attention from the research community. Researchers can explore this area  through data \cite{zhao2021data,you2020graph}, model design \cite{li2022ood,fan2023generalizing}, and training strategies \cite{li2022learning,hu2019strategies}.   Time-series OOD has also garnered considerable interest, particularly in forecasting \cite{wen2024onenet}, classification \cite{lu2022out}, and detection \cite{lu2024diversify} tasks. Some works approach this from an invariance perspective \cite{liu2024time}, while others focus on normalization techniques \cite{kim2021reversible,liu2022non}. 
	
	\item[\ding{117}] \textbf{Developing spatiotemporal OOD theory.} Although some recent pilot studies have proposed preliminary spatiotemporal OOD methods, such as \cite{xia2024deciphering}, which employs disentanglement blocks for backdoor adjustment by separating temporal contexts from input data, and frontdoor adjustment with edge-level convolutions to model causal chain effects, as well as \cite{zhou2023maintaining}, which redefines invariant learning by capturing stable, trainable weights, these approaches largely rely on a separation strategy for temporal and spatial OOD. As indicated by the comparative results mentioned earlier, their failures are likely attributable to this spatiotemporal separation. Therefore, a truly comprehensive and novel spatiotemporal OOD theory remains to be developed.
	
	\item[\ding{117}] \textbf{Leveraging large language models (LLMs)  for predicting spatiotemporal OOD.} LLMs has recently garnered significant attention due to their unprecedented performance in tasks such as text comprehension and reasoning. These models, including open-source LLMs like Llama \cite{touvron2023llama,touvron2023llama2,dubey2024llama}, have demonstrated the potential to extend from intelligent algorithms to artificial general intelligence. Researchers are actively exploring their application across various domains to enhance transfer learning through the domain knowledge embedded in these models. UrbanGPT \cite{li2024urbangpt}, for instance, has shown promising zero-shot generalization capabilities in spatiotemporal task. We hypothesize that LLMs are inherently well-suited for handling urban OOD scenarios, as they can be applied to new demands or event-driven situations, such as COVID-19 or typhoons, by leveraging their deep understanding and reasoning abilities.

	\item[\ding{117}] \textbf{Developing simple yet effective methods.} From the analysis of \tableref{tbl:res}, it is evident that while recent methods demonstrate increasing accuracy, they also exhibit growing model complexity. Although models such as STAEformer and D$^2$STGNN perform well on in-distribution data, the dynamic nature of urban development and the evolving spatiotemporal attributes indicate that simpler methods, such as STID and MLP, often exhibit superior generalization. We find that the primary distinction lies in the fact that current state-of-the-art methods introduce excessive inductive biases, causing the models to learn an abundance of spurious relationships.
	Therefore, there is an urgent need to develop straightforward and effective spatiotemporal prediction models that can be practically implemented and deployed in real-world applications to address spatiotemporal shifts.
\end{itemize}


\bibliographystyle{IEEEtran}
\bibliography{reference}

\begin{thebibliography}{10}
\providecommand{\url}[1]{#1}
\csname url@samestyle\endcsname
\providecommand{\newblock}{\relax}
\providecommand{\bibinfo}[2]{#2}
\providecommand{\BIBentrySTDinterwordspacing}{\spaceskip=0pt\relax}
\providecommand{\BIBentryALTinterwordstretchfactor}{4}
\providecommand{\BIBentryALTinterwordspacing}{\spaceskip=\fontdimen2\font plus
\BIBentryALTinterwordstretchfactor\fontdimen3\font minus
  \fontdimen4\font\relax}
\providecommand{\BIBforeignlanguage}[2]{{%
\expandafter\ifx\csname l@#1\endcsname\relax
\typeout{** WARNING: IEEEtran.bst: No hyphenation pattern has been}%
\typeout{** loaded for the language `#1'. Using the pattern for}%
\typeout{** the default language instead.}%
\else
\language=\csname l@#1\endcsname
\fi
#2}}
\providecommand{\BIBdecl}{\relax}
\BIBdecl

\bibitem{torralba2011unbiased}
A.~Torralba and A.~Efros, ``Unbiased look at dataset bias,'' \emph{CVPR}, 2011.

\bibitem{dai2018dark}
D.~Dai and L.~Van~Gool, ``Dark model adaptation: Semantic image segmentation
  from daytime to nighttime,'' \emph{ITSC}, 2018.

\bibitem{volk2019towards}
G.~Volk, S.~M{\"u}ller, A.~von Bernuth, D.~Hospach, and O.~Bringmann, ``Towards
  robust cnn-based object detection through augmentation with synthetic rain
  variations,'' \emph{ITSC}, 2019.

\bibitem{alcorn2019strike}
M.~A. Alcorn, Q.~Li, Z.~Gong, C.~Wang, L.~Mai, W.-S. Ku, and A.~Nguyen,
  ``Strike (with) a pose: Neural networks are easily fooled by strange poses of
  familiar objects,'' \emph{CVPR}, 2019.

\bibitem{zhang2020curb}
Y.~Zhang, Y.~Li, X.~Zhou, X.~Kong, and J.~Luo, ``Curb-gan: Conditional urban
  traffic estimation through spatio-temporal generative adversarial networks,''
  in \emph{Proceedings of the 26th ACM SIGKDD International Conference on
  Knowledge Discovery \& Data Mining}, 2020, pp. 842--852.

\bibitem{li2018diffusion}
Y.~Li, R.~Yu, C.~Shahabi, and Y.~Liu, ``Diffusion convolutional recurrent
  neural network: Data-driven traffic forecasting,'' in \emph{International
  Conference on Learning Representations}, 2018.

\bibitem{yu2018spatio}
B.~Yu, H.~Yin, and Z.~Zhu, ``Spatio-temporal graph convolutional networks: A
  deep learning framework for traffic forecasting,'' in \emph{Proceedings of
  International Joint Conference on Artificial Intelligence}, 2018, pp.
  3634--3640.

\bibitem{liu2024largest}
X.~Liu, Y.~Xia, Y.~Liang, J.~Hu, Y.~Wang, L.~Bai, C.~Huang, Z.~Liu, B.~Hooi,
  and R.~Zimmermann, ``Largest: A benchmark dataset for large-scale traffic
  forecasting,'' \emph{Advances in Neural Information Processing Systems},
  vol.~36, 2024.

\bibitem{liu2023spatio}
H.~Liu, Z.~Dong, R.~Jiang, J.~Deng, J.~Deng, Q.~Chen, and X.~Song,
  ``Spatio-temporal adaptive embedding makes vanilla transformer sota for
  traffic forecasting,'' in \emph{Proceedings of the 32nd ACM international
  conference on information and knowledge management}, 2023, pp. 4125--4129.

\bibitem{bai2020adaptive}
L.~Bai, L.~Yao, C.~Li, X.~Wang, and C.~Wang, ``Adaptive graph convolutional
  recurrent network for traffic forecasting,'' in \emph{Proceedings of Advances
  in Neural Information Processing Systems}, 2020, pp. 17\,804--17\,815.

\bibitem{pearl2009causality}
J.~Pearl, \emph{Causality}.\hskip 1em plus 0.5em minus 0.4em\relax Cambridge
  university press, 2009.

\bibitem{tobler1970computer}
W.~R. Tobler, ``A computer movie simulating urban growth in the detroit
  region,'' \emph{Economic geography}, vol.~46, no. sup1, pp. 234--240, 1970.

\bibitem{wu2019graph}
Z.~Wu, S.~Pan, G.~Long, J.~Jiang, and C.~Zhang, ``Graph wavenet for deep
  spatial-temporal graph modeling,'' in \emph{Proceedings of International
  Joint Conference on Artificial Intelligence}, 2019, pp. 1907--1913.

\bibitem{fang2021spatial}
Z.~Fang, Q.~Long, G.~Song, and K.~Xie, ``Spatial-temporal graph ode networks
  for traffic flow forecasting,'' in \emph{Proceedings of the 27th ACM SIGKDD
  Conference on Knowledge Discovery and Data Mining}, 2021, pp. 364--373.

\bibitem{shao2022spatial}
Z.~Shao, Z.~Zhang, F.~Wang, W.~Wei, and Y.~Xu, ``Spatial-temporal identity: A
  simple yet effective baseline for multivariate time series forecasting,'' in
  \emph{Proceedings of the 31st ACM International Conference on Information \&
  Knowledge Management}, 2022, pp. 4454--4458.

\bibitem{asgari2022masktune}
S.~Asgari, A.~Khani, F.~Khani, A.~Gholami, L.~Tran, A.~Mahdavi~Amiri, and
  G.~Hamarneh, ``Masktune: Mitigating spurious correlations by forcing to
  explore,'' \emph{Advances in Neural Information Processing Systems}, vol.~35,
  pp. 23\,284--23\,296, 2022.

\bibitem{xia2024deciphering}
Y.~Xia, Y.~Liang, H.~Wen, X.~Liu, K.~Wang, Z.~Zhou, and R.~Zimmermann,
  ``Deciphering spatio-temporal graph forecasting: A causal lens and
  treatment,'' \emph{Advances in Neural Information Processing Systems},
  vol.~36, 2024.

\bibitem{zhou2023maintaining}
Z.~Zhou, Q.~Huang, K.~Yang, K.~Wang, X.~Wang, Y.~Zhang, Y.~Liang, and Y.~Wang,
  ``Maintaining the status quo: Capturing invariant relations for ood
  spatiotemporal learning,'' in \emph{Proceedings of the 29th ACM SIGKDD
  Conference on Knowledge Discovery and Data Mining}, 2023, pp. 3603--3614.

\bibitem{wang2024stone}
B.~Wang, J.~Ma, P.~Wang, X.~Wang, Y.~Zhang, Z.~Zhou, and Y.~Wang, ``Stone: A
  spatio-temporal ood learning framework kills both spatial and temporal
  shifts,'' in \emph{Proceedings of the 30th ACM SIGKDD Conference on Knowledge
  Discovery and Data Mining}, 2024, pp. 2948--2959.

\bibitem{guo2019attention}
S.~Guo, Y.~Lin, N.~Feng, C.~Song, and H.~Wan, ``Attention based
  spatial-temporal graph convolutional networks for traffic flow forecasting,''
  in \emph{Proceedings of the AAAI Conference on Artificial Intelligence},
  vol.~33, no.~01, 2019, pp. 922--929.

\bibitem{sun2020predicting}
J.~Sun, J.~Zhang, Q.~Li, X.~Yi, Y.~Liang, and Y.~Zheng, ``Predicting citywide
  crowd flows in irregular regions using multi-view graph convolutional
  networks,'' \emph{IEEE Transactions on Knowledge and Data Engineering},
  vol.~34, no.~5, pp. 2348--2359, 2020.

\bibitem{chai2018bike}
D.~Chai, L.~Wang, and Q.~Yang, ``Bike flow prediction with multi-graph
  convolutional networks,'' in \emph{Proceedings of the 26th ACM SIGSPATIAL
  international conference on advances in geographic information systems},
  2018, pp. 397--400.

\bibitem{lin2018exploiting}
Y.~Lin, N.~Mago, Y.~Gao, Y.~Li, Y.-Y. Chiang, C.~Shahabi, and J.~L. Ambite,
  ``Exploiting spatiotemporal patterns for accurate air quality forecasting
  using deep learning,'' in \emph{Proceedings of the 26th ACM SIGSPATIAL
  international conference on advances in geographic information systems},
  2018, pp. 359--368.

\bibitem{lin2022conditional}
H.~Lin, Z.~Gao, Y.~Xu, L.~Wu, L.~Li, and S.~Z. Li, ``Conditional local
  convolution for spatio-temporal meteorological forecasting,'' in
  \emph{Proceedings of the AAAI conference on artificial intelligence},
  vol.~36, no.~7, 2022, pp. 7470--7478.

\bibitem{xia2021spatial}
L.~Xia, C.~Huang, Y.~Xu, P.~Dai, L.~Bo, X.~Zhang, and T.~Chen,
  ``Spatial-temporal sequential hypergraph network for crime prediction with
  dynamic multiplex relation learning.'' in \emph{IJCAI}, 2021, pp. 1631--1637.

\bibitem{yang2022spatio}
X.~Yang, F.~Zhang, P.~Sun, X.~Li, Z.~Du, and R.~Liu, ``A spatio-temporal
  graph-guided convolutional lstm for tropical cyclones precipitation
  nowcasting,'' \emph{Applied Soft Computing}, vol. 124, p. 109003, 2022.

\bibitem{wang2022causalgnn}
L.~Wang, A.~Adiga, J.~Chen, A.~Sadilek, S.~Venkatramanan, and M.~Marathe,
  ``Causalgnn: Causal-based graph neural networks for spatio-temporal epidemic
  forecasting,'' in \emph{Proceedings of the AAAI conference on artificial
  intelligence}, vol.~36, no.~11, 2022, pp. 12\,191--12\,199.

\bibitem{deng2019graph}
S.~Deng, S.~Wang, H.~Rangwala, L.~Wang, and Y.~Ning, ``Graph message passing
  with cross-location attentions for long-term ili prediction,'' \emph{arXiv
  preprint arXiv:1912.10202}, 2019.

\bibitem{shao2023exploring}
Z.~Shao, F.~Wang, Y.~Xu, W.~Wei, C.~Yu, Z.~Zhang, D.~Yao, G.~Jin, X.~Cao,
  G.~Cong \emph{et~al.}, ``Exploring progress in multivariate time series
  forecasting: Comprehensive benchmarking and heterogeneity analysis,''
  \emph{arXiv preprint arXiv:2310.06119}, 2023.

\bibitem{rasp2023weatherbench}
S.~Rasp, S.~Hoyer, A.~Merose, I.~Langmore, P.~Battaglia, T.~Russel,
  A.~Sanchez-Gonzalez, V.~Yang, R.~Carver, S.~Agrawal, M.~Chantry, Z.~B.
  Bouallegue, P.~Dueben, C.~Bromberg, J.~Sisk, L.~Barrington, A.~Bell, and
  F.~Sha, ``Weatherbench 2: A benchmark for the next generation of data-driven
  global weather models,'' 2023.

\bibitem{cao2020spectral}
D.~Cao, Y.~Wang, J.~Duan, C.~Zhang, X.~Zhu, C.~Huang, Y.~Tong, B.~Xu, J.~Bai,
  J.~Tong \emph{et~al.}, ``Spectral temporal graph neural network for
  multivariate time-series forecasting,'' in \emph{Proceedings of Advances in
  Neural Information Processing Systems}, 2020, pp. 17\,766--17\,778.

\bibitem{wang2023easy}
H.~Wang, J.~Chen, T.~Pan, Z.~Fan, X.~Song, R.~Jiang, L.~Zhang, Y.~Xie, Z.~Wang,
  and B.~Zhang, ``Easy begun is half done: spatial-temporal graph modeling with
  st-curriculum dropout,'' in \emph{Proceedings of the AAAI Conference on
  Artificial Intelligence}, vol.~37, no.~4, 2023, pp. 4668--4675.

\bibitem{wang2023multi}
H.~Wang, Z.~Zhang, Z.~Fan, J.~Chen, L.~Zhang, R.~Shibasaki, and X.~Song,
  ``Multi-task weakly supervised learning for origin--destination travel time
  estimation,'' \emph{IEEE Transactions on Knowledge and Data Engineering},
  vol.~35, no.~11, pp. 11\,628--11\,641, 2023.

\bibitem{wang2022st}
H.~Wang, J.~Chen, Z.~Fan, Z.~Zhang, Z.~Cai, and X.~Song, ``St-expertnet: A deep
  expert framework for traffic prediction,'' \emph{IEEE Transactions on
  Knowledge and Data Engineering}, 2022.

\bibitem{pan2019urban}
Z.~Pan, Y.~Liang, W.~Wang, Y.~Yu, Y.~Zheng, and J.~Zhang, ``Urban traffic
  prediction from spatio-temporal data using deep meta learning,'' in
  \emph{Proceedings of the 25th ACM SIGKDD Conference on Knowledge Discovery
  and Data Mining}, 2019, pp. 1720--1730.

\bibitem{li2021dynamic}
F.~Li, J.~Feng, H.~Yan, G.~Jin, F.~Yang, F.~Sun, D.~Jin, and Y.~Li, ``Dynamic
  graph convolutional recurrent network for traffic prediction: Benchmark and
  solution,'' \emph{ACM Transactions on Knowledge Discovery from Data}, pp.
  1--21, 2021.

\bibitem{han2021dynamic}
L.~Han, B.~Du, L.~Sun, Y.~Fu, Y.~Lv, and H.~Xiong, ``Dynamic and multi-faceted
  spatio-temporal deep learning for traffic speed forecasting,'' in
  \emph{Proceedings of the 27th ACM SIGKDD Conference on Knowledge Discovery
  and Data Mining}, 2021, pp. 547--555.

\bibitem{choi2022graph}
J.~Choi, H.~Choi, J.~Hwang, and N.~Park, ``Graph neural controlled differential
  equations for traffic forecasting,'' in \emph{Proceedings of the AAAI
  Conference on Artificial Intelligence}, 2022, pp. 6367--6374.

\bibitem{lan2022dstagnn}
S.~Lan, Y.~Ma, W.~Huang, W.~Wang, H.~Yang, and P.~Li, ``Dstagnn: Dynamic
  spatial-temporal aware graph neural network for traffic flow forecasting,''
  in \emph{International conference on machine learning}.\hskip 1em plus 0.5em
  minus 0.4em\relax PMLR, 2022, pp. 11\,906--11\,917.

\bibitem{shao2022decoupled}
Z.~Shao, Z.~Zhang, W.~Wei, F.~Wang, Y.~Xu, X.~Cao, and C.~S. Jensen,
  ``Decoupled dynamic spatial-temporal graph neural network for traffic
  forecasting,'' in \emph{Proceedings of the VLDB Endowment}, 2022, pp.
  2733--2746.

\bibitem{du2021adarnn}
Y.~Du, J.~Wang, W.~Feng, S.~Pan, T.~Qin, R.~Xu, and C.~Wang, ``Adarnn: Adaptive
  learning and forecasting of time series,'' in \emph{Proceedings of the 30th
  ACM international conference on information \& knowledge management}, 2021,
  pp. 402--411.

\bibitem{yao2022wild}
H.~Yao, C.~Choi, B.~Cao, Y.~Lee, P.~W.~W. Koh, and C.~Finn, ``Wild-time: A
  benchmark of in-the-wild distribution shift over time,'' \emph{Advances in
  Neural Information Processing Systems}, vol.~35, pp. 10\,309--10\,324, 2022.

\bibitem{deng2023spatio}
P.~Deng, Y.~Zhao, J.~Liu, X.~Jia, and M.~Wang, ``Spatio-temporal neural
  structural causal models for bike flow prediction,'' in \emph{Proceedings of
  the AAAI conference on artificial intelligence}, vol.~37, no.~4, 2023, pp.
  4242--4249.

\bibitem{wu2020connecting}
Z.~Wu, S.~Pan, G.~Long, J.~Jiang, X.~Chang, and C.~Zhang, ``Connecting the
  dots: Multivariate time series forecasting with graph neural networks,'' in
  \emph{Proceedings of the 26th ACM SIGKDD international conference on
  knowledge discovery \& data mining}, 2020, pp. 753--763.

\bibitem{jiang2021dl}
R.~Jiang, D.~Yin, Z.~Wang, Y.~Wang, J.~Deng, H.~Liu, Z.~Cai, J.~Deng, X.~Song,
  and R.~Shibasaki, ``Dl-traff: Survey and benchmark of deep learning models
  for urban traffic prediction,'' in \emph{Proceedings of the 30th ACM
  international conference on information \& knowledge management}, 2021, pp.
  4515--4525.

\bibitem{kendall1938new}
M.~G. Kendall, ``A new measure of rank correlation,'' \emph{Biometrika},
  vol.~30, no. 1/2, pp. 81--93, 1938.

\bibitem{edwards2023graphing}
N.~D. Edwards, E.~de~Jong, and S.~T. Ferguson, ``Graphing methods for kendall's
  $\{$$\backslash$tau$\}$,'' \emph{arXiv preprint arXiv:2308.08466}, 2023.

\bibitem{muller2007dynamic}
M.~M{\"u}ller, ``Dynamic time warping,'' \emph{Information retrieval for music
  and motion}, pp. 69--84, 2007.

\bibitem{santana2023covid}
C.~Santana, F.~Botta, H.~Barbosa, F.~Privitera, R.~Menezes, and R.~Di~Clemente,
  ``Covid-19 is linked to changes in the time--space dimension of human
  mobility,'' \emph{Nature Human Behaviour}, vol.~7, no.~10, pp. 1729--1739,
  2023.

\bibitem{Jackson}
J.~Fordham, ``How much have 311 service requests changed during the covid-19
  pandemic in new york city?'' [Online], 2020, \url{ https://reurl.cc/xvLVQ4}.

\bibitem{hochreiter1997long}
S.~Hochreiter, J.~urgen Schmidhuber, and C.~Elvezia, ``Long short-term
  memory,'' \emph{Neural Computation}, vol.~9, no.~8, pp. 1735--1780, 1997.

\bibitem{van2016wavenet}
A.~Van Den~Oord, S.~Dieleman, H.~Zen, K.~Simonyan, O.~Vinyals, A.~Graves,
  N.~Kalchbrenner, A.~Senior, K.~Kavukcuoglu \emph{et~al.}, ``Wavenet: A
  generative model for raw audio,'' \emph{arXiv preprint arXiv:1609.03499},
  vol.~12, 2016.

\bibitem{xu2020spatial}
M.~Xu, W.~Dai, C.~Liu, X.~Gao, W.~Lin, G.-J. Qi, and H.~Xiong,
  ``Spatial-temporal transformer networks for traffic flow forecasting,''
  \emph{arXiv preprint arXiv:2001.02908}, 2020.

\bibitem{oord2016WaveNET}
A.~v.~d. Oord, S.~Dieleman, H.~Zen, K.~Simonyan, O.~Vinyals, A.~Graves,
  N.~Kalchbrenner, A.~Senior, and K.~Kavukcuoglu, ``Wavenet: A generative model
  for raw audio,'' \emph{arXiv preprint arXiv:1609.03499}, 2016.

\bibitem{kipf2016semi}
T.~N. Kipf and M.~Welling, ``Semi-supervised classification with graph
  convolutional networks,'' \emph{ICLR}, 2017.

\bibitem{vaswani2017attention}
A.~Vaswani, N.~Shazeer, N.~Parmar, J.~Uszkoreit, L.~Jones, A.~N. Gomez,
  L.~Kaiser, and I.~Polosukhin, ``Attention is all you need,'' in
  \emph{NeurIPS}, 2017.

\bibitem{srinivasan2019equivalence}
B.~Srinivasan and B.~Ribeiro, ``On the equivalence between positional node
  embeddings and structural graph representations,'' \emph{arXiv preprint
  arXiv:1910.00452}, 2019.

\bibitem{rahman2019fairwalk}
T.~Rahman, B.~Surma, M.~Backes, and Y.~Zhang, ``Fairwalk: Towards fair graph
  embedding,'' 2019.

\bibitem{xu2019spatio}
D.~Xu, W.~Cheng, D.~Luo, X.~Liu, and X.~Zhang, ``Spatio-temporal attentive rnn
  for node classification in temporal attributed graphs.'' in \emph{IJCAI},
  2019, pp. 3947--3953.

\bibitem{krueger2021out}
D.~Krueger, E.~Caballero, J.-H. Jacobsen, A.~Zhang, J.~Binas, D.~Zhang,
  R.~Le~Priol, and A.~Courville, ``Out-of-distribution generalization via risk
  extrapolation (rex),'' in \emph{ICML}, 2021.

\bibitem{li2022out}
H.~Li, X.~Wang, Z.~Zhang, and W.~Zhu, ``Out-of-distribution generalization on
  graphs: A survey,'' \emph{arXiv preprint arXiv:2202.07987}, 2022.

\bibitem{zhao2021data}
T.~Zhao, Y.~Liu, L.~Neves, O.~Woodford, M.~Jiang, and N.~Shah, ``Data
  augmentation for graph neural networks,'' in \emph{Proceedings of the aaai
  conference on artificial intelligence}, vol.~35, no.~12, 2021, pp.
  11\,015--11\,023.

\bibitem{you2020graph}
Y.~You, T.~Chen, Y.~Sui, T.~Chen, Z.~Wang, and Y.~Shen, ``Graph contrastive
  learning with augmentations,'' \emph{Advances in neural information
  processing systems}, vol.~33, pp. 5812--5823, 2020.

\bibitem{li2022ood}
H.~Li, X.~Wang, Z.~Zhang, and W.~Zhu, ``Ood-gnn: Out-of-distribution
  generalized graph neural network,'' \emph{IEEE Transactions on Knowledge and
  Data Engineering}, vol.~35, no.~7, pp. 7328--7340, 2022.

\bibitem{fan2023generalizing}
S.~Fan, X.~Wang, C.~Shi, P.~Cui, and B.~Wang, ``Generalizing graph neural
  networks on out-of-distribution graphs,'' \emph{IEEE Transactions on Pattern
  Analysis and Machine Intelligence}, 2023.

\bibitem{li2022learning}
H.~Li, Z.~Zhang, X.~Wang, and W.~Zhu, ``Learning invariant graph
  representations for out-of-distribution generalization,'' \emph{Advances in
  Neural Information Processing Systems}, vol.~35, pp. 11\,828--11\,841, 2022.

\bibitem{hu2019strategies}
W.~Hu, B.~Liu, J.~Gomes, M.~Zitnik, P.~Liang, V.~Pande, and J.~Leskovec,
  ``Strategies for pre-training graph neural networks,'' \emph{arXiv preprint
  arXiv:1905.12265}, 2019.

\bibitem{wen2024onenet}
Q.~Wen, W.~Chen, L.~Sun, Z.~Zhang, L.~Wang, R.~Jin, T.~Tan \emph{et~al.},
  ``Onenet: Enhancing time series forecasting models under concept drift by
  online ensembling,'' \emph{Advances in Neural Information Processing
  Systems}, vol.~36, 2024.

\bibitem{lu2022out}
W.~Lu, J.~Wang, X.~Sun, Y.~Chen, and X.~Xie, ``Out-of-distribution
  representation learning for time series classification,'' \emph{arXiv
  preprint arXiv:2209.07027}, 2022.

\bibitem{lu2024diversify}
W.~Lu, J.~Wang, X.~Sun, Y.~Chen, X.~Ji, Q.~Yang, and X.~Xie, ``Diversify: A
  general framework for time series out-of-distribution detection and
  generalization,'' \emph{IEEE Transactions on Pattern Analysis and Machine
  Intelligence}, 2024.

\bibitem{liu2024time}
H.~Liu, H.~Kamarthi, L.~Kong, Z.~Zhao, C.~Zhang, and B.~A. Prakash,
  ``Time-series forecasting for out-of-distribution generalization using
  invariant learning,'' \emph{arXiv preprint arXiv:2406.09130}, 2024.

\bibitem{kim2021reversible}
T.~Kim, J.~Kim, Y.~Tae, C.~Park, J.-H. Choi, and J.~Choo, ``Reversible instance
  normalization for accurate time-series forecasting against distribution
  shift,'' in \emph{International Conference on Learning Representations},
  2021.

\bibitem{liu2022non}
Y.~Liu, H.~Wu, J.~Wang, and M.~Long, ``Non-stationary transformers: Exploring
  the stationarity in time series forecasting,'' \emph{Advances in Neural
  Information Processing Systems}, vol.~35, pp. 9881--9893, 2022.

\bibitem{touvron2023llama}
H.~Touvron, T.~Lavril, G.~Izacard, X.~Martinet, M.-A. Lachaux, T.~Lacroix,
  B.~Rozi{\`e}re, N.~Goyal, E.~Hambro, F.~Azhar \emph{et~al.}, ``Llama: Open
  and efficient foundation language models,'' \emph{arXiv preprint
  arXiv:2302.13971}, 2023.

\bibitem{touvron2023llama2}
H.~Touvron, L.~Martin, K.~Stone, P.~Albert, A.~Almahairi, Y.~Babaei,
  N.~Bashlykov, S.~Batra, P.~Bhargava, S.~Bhosale \emph{et~al.}, ``Llama 2:
  Open foundation and fine-tuned chat models,'' \emph{arXiv preprint
  arXiv:2307.09288}, 2023.

\bibitem{dubey2024llama}
A.~Dubey, A.~Jauhri, A.~Pandey, A.~Kadian, A.~Al-Dahle, A.~Letman, A.~Mathur,
  A.~Schelten, A.~Yang, A.~Fan \emph{et~al.}, ``The llama 3 herd of models,''
  \emph{arXiv preprint arXiv:2407.21783}, 2024.

\bibitem{li2024urbangpt}
Z.~Li, L.~Xia, J.~Tang, Y.~Xu, L.~Shi, L.~Xia, D.~Yin, and C.~Huang,
  ``Urbangpt: Spatio-temporal large language models,'' in \emph{Proceedings of
  the 30th ACM SIGKDD Conference on Knowledge Discovery and Data Mining}, 2024,
  pp. 5351--5362.

\end{thebibliography}
\begin{IEEEbiography}[{\includegraphics[width=1in,height=1.25in,clip,keepaspectratio]{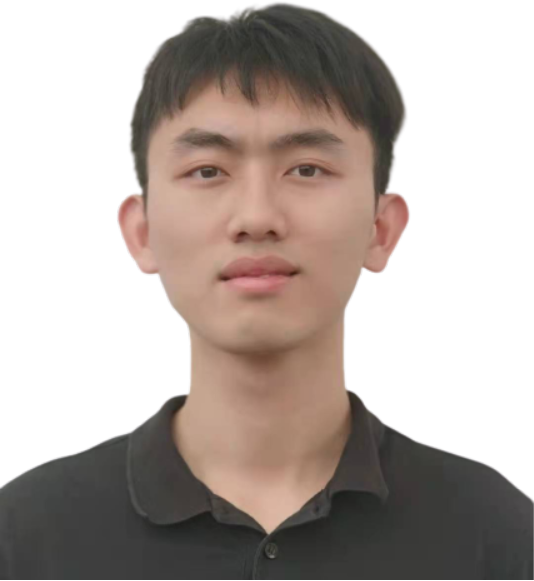}}]{Hongjun Wang} is working toward the PhD degree in the Department of Mechano-Informatics at The University of Tokyo. He received his M.S. degree in computer science and technology from Southern University of Science and Technology, China. He received his B.E. degree from the Nanjing University of Posts and Telecommunications, China, in 2019. His research interests are broadly in machine learning, with a focus on urban computing, explainable AI, data mining, and data visualization.
\end{IEEEbiography}
\vspace{1ex}
\begin{IEEEbiography}[{\includegraphics[width=1in,height=1.25in,clip,keepaspectratio]{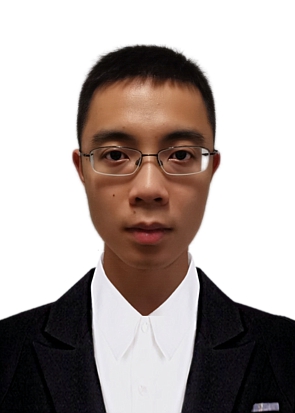}}]{Jiyuan Chen} is working towards his PhD degree at The Hong Kong Polytechnic University. He received his B.S. degree in Computer Science and Technology from Southern University of Science and Technology, China. His major research fields include artificial intelligence, deep learning, urban computing, and data mining.
\end{IEEEbiography}
\vspace{1ex}
\begin{IEEEbiography}[{\includegraphics[width=1in,height=1.25in,clip,keepaspectratio]{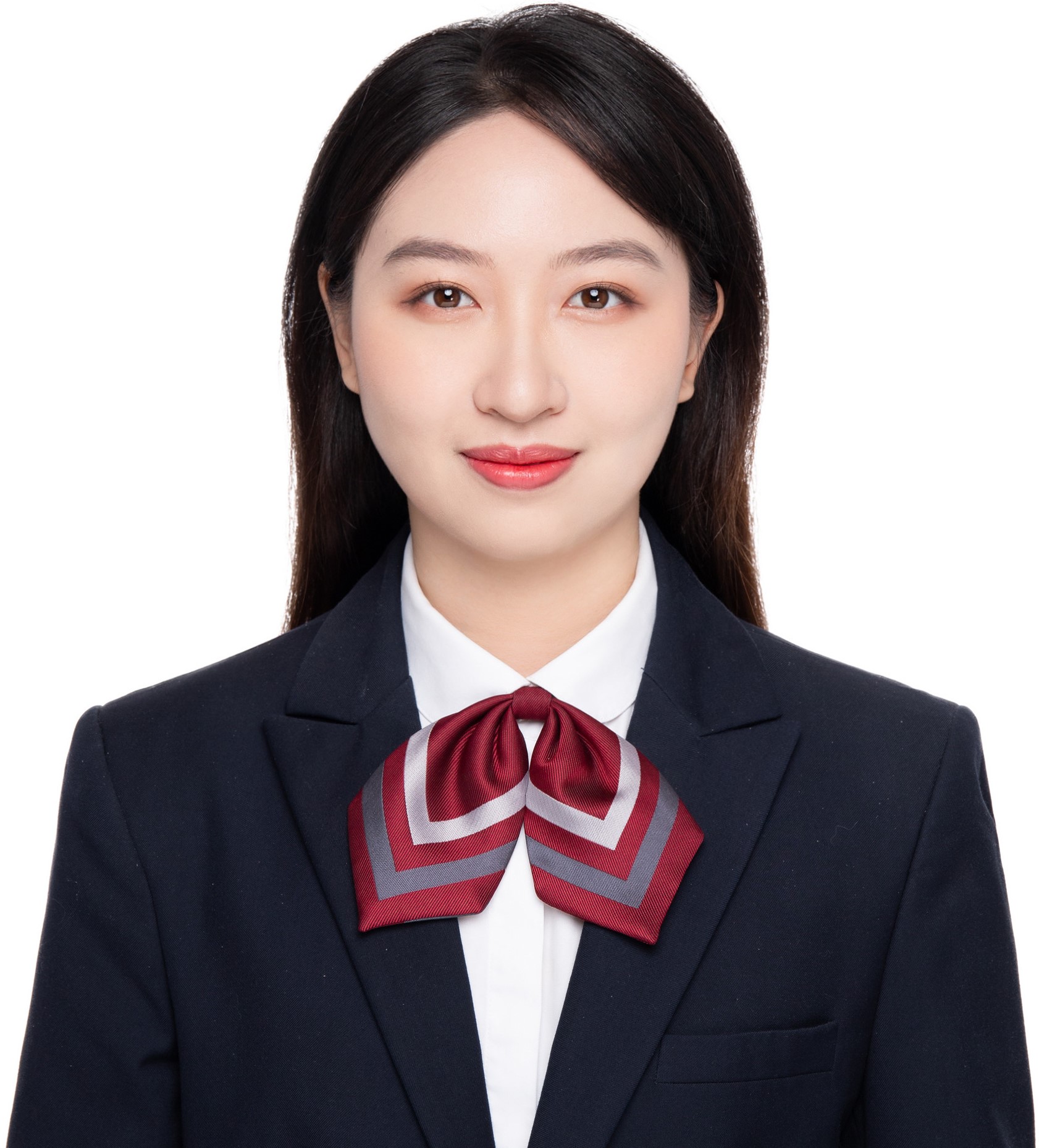}}]{Tong Pan} received a B.S. degree in Physics from East China Normal University, China, in 2019, and a Ph.D. degree in Physics from the Chinese University of Hong Kong, China, in 2024. From 2024, she has been a postdoctal researcher at Southern University of Science and Technology, China. Her research interests include data analysis, machine learning and AI for science. 
\end{IEEEbiography}
\vspace{1ex}
\begin{IEEEbiography}[{\includegraphics[width=1in,height=1.25in,clip,keepaspectratio]{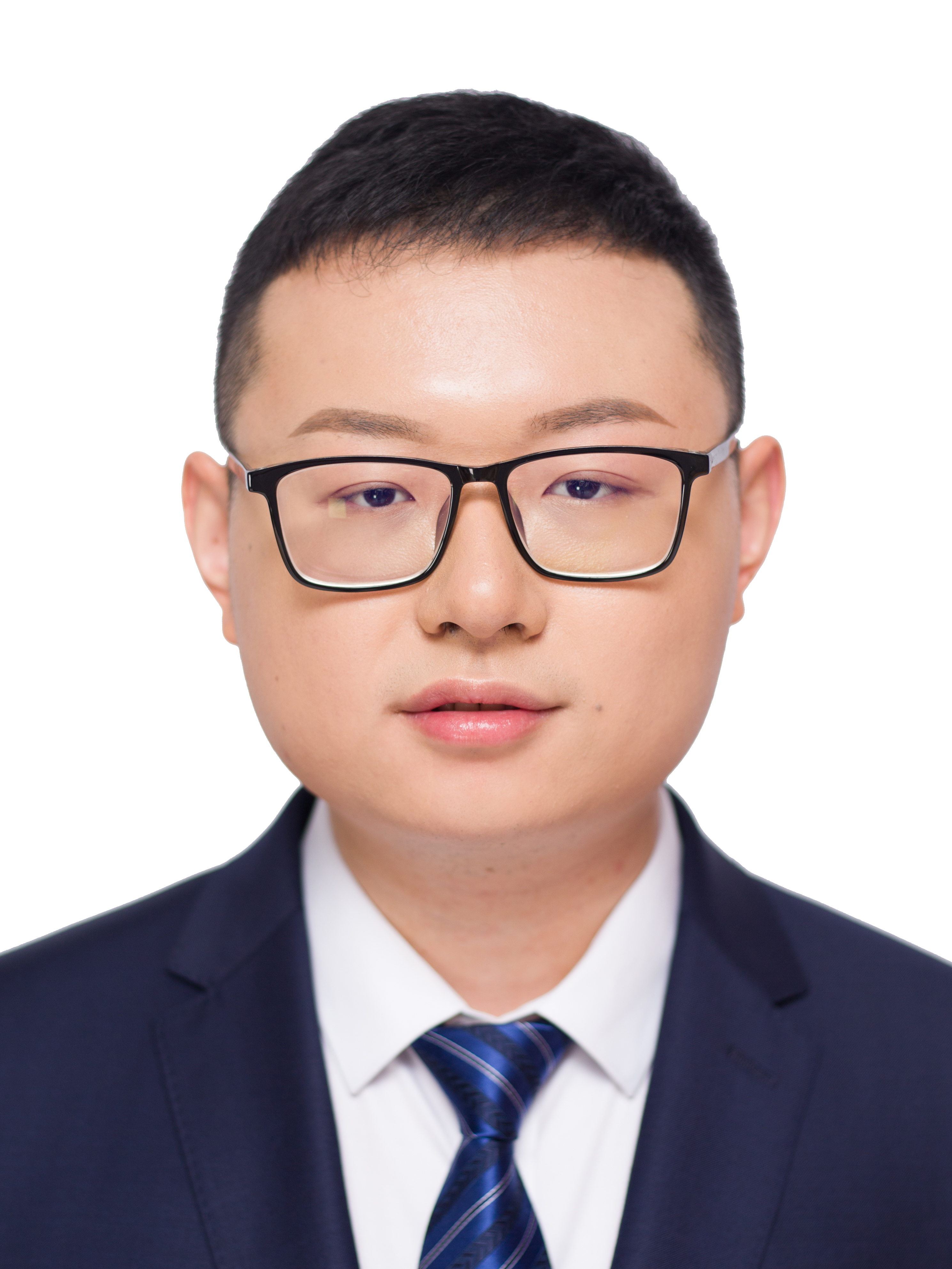}}]{Zheng Dong}  received his B.E. degree in computer science and technology from Southern University of Science and Technology (SUSTech) in 2022. He is currently persuing a M.S. degree in the Department of Computer Science and Engineering, SUSTech. His research interests include deep learning and spatio-temporal data mining.
\end{IEEEbiography}

\vspace{1ex}
\begin{IEEEbiography}[{\includegraphics[width=1in,height=1.25in,clip,keepaspectratio]{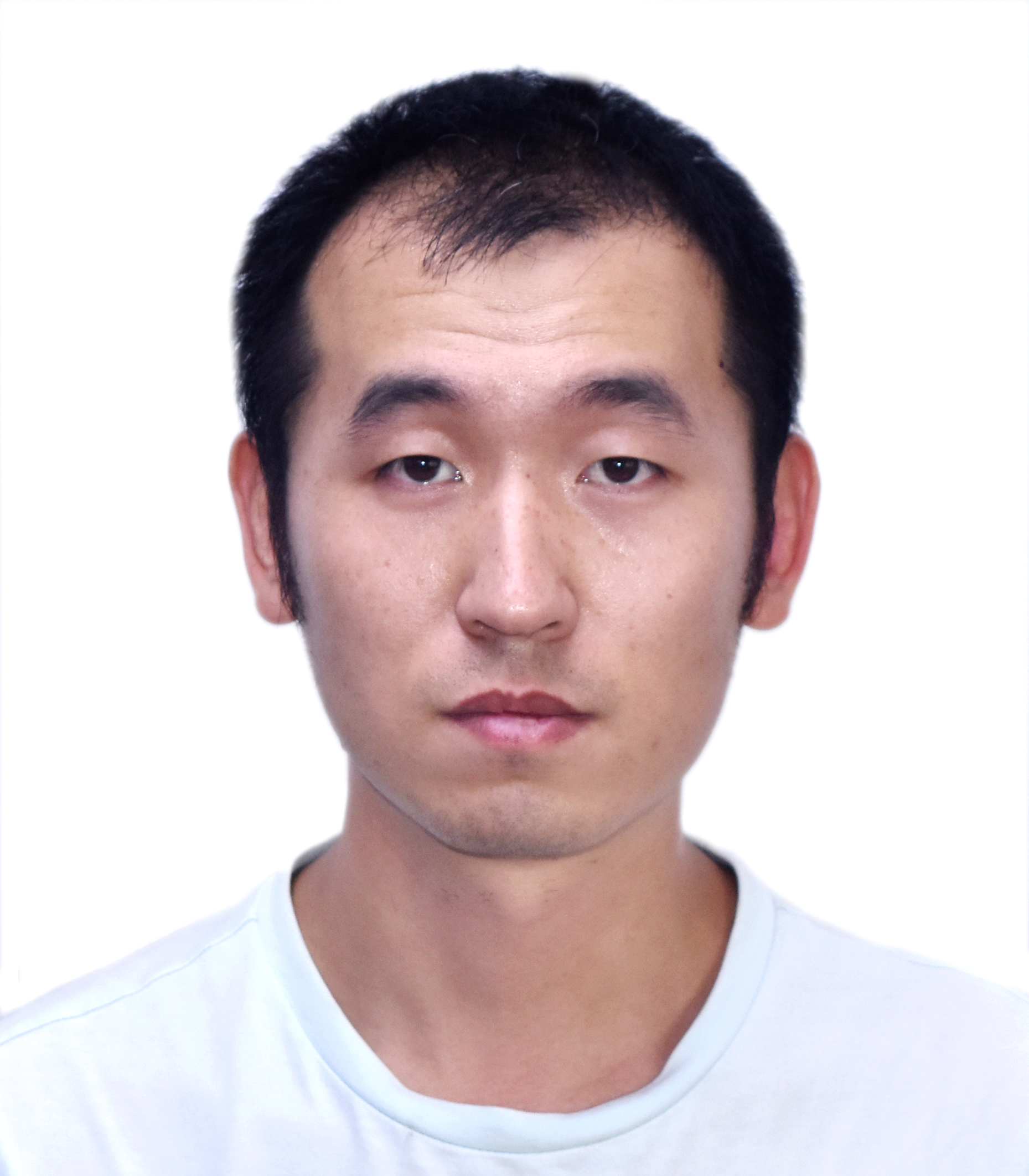}}]{Lingyu Zhang}   joined Baidu in 2012 as a search strategy algorithm research and development engineer. He joined Didi in 2013 and served as senior algorithm engineer, technical director of taxi strategy algorithm direction, and technical expert of strategy model department. Currently a researcher at Didi AI Labs, he used machine learning and big data technology to design and lead the implementation of multiple company-level intelligent system engines during his work at Didi, such as the order distribution system based on combination optimization, and the capacity based on density clustering and global optimization. Scheduling engine, traffic guidance and personalized recommendation engine, "Guess where you are going" personalized destination recommendation system, etc. Participated in the company's dozens of international and domestic core technology innovation patent research and development, application, good at using mathematical modeling, business model abstraction, machine learning, etc. to solve practical business problems. He has won honorary titles such as Beijing Invention and Innovation Patent Gold Award and QCon Star Lecturer, and his research results have been included in top international conferences related to artificial intelligence and data mining such as KDD, SIGIR, AAAI, and CIKM.
\end{IEEEbiography}
\vspace{1ex}
\begin{IEEEbiography}[{\includegraphics[width=1in,height=1.25in,clip,keepaspectratio]{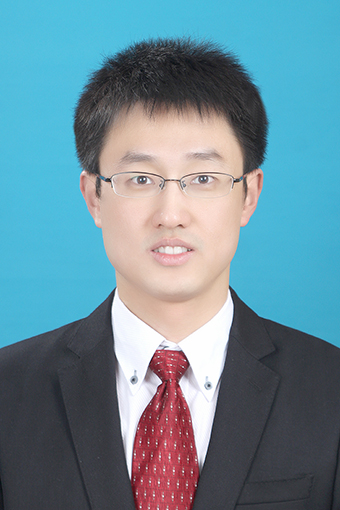}}]{Renhe Jiang} received a B.S. degree in software engineering from the Dalian University of Technology, China, in 2012, a M.S. degree in information science from Nagoya University, Japan, in 2015, and a Ph.D. degree in civil engineering from The University of Tokyo, Japan, in 2019. From 2019, he has been an Assistant Professor at the Information Technology Center, The University of Tokyo. His research interests include ubiquitous computing, deep learning, and spatio-temporal data analysis.
\end{IEEEbiography}
\vspace{1ex}
\begin{IEEEbiography}[{\includegraphics[width=1in,height=1.25in,clip,keepaspectratio]{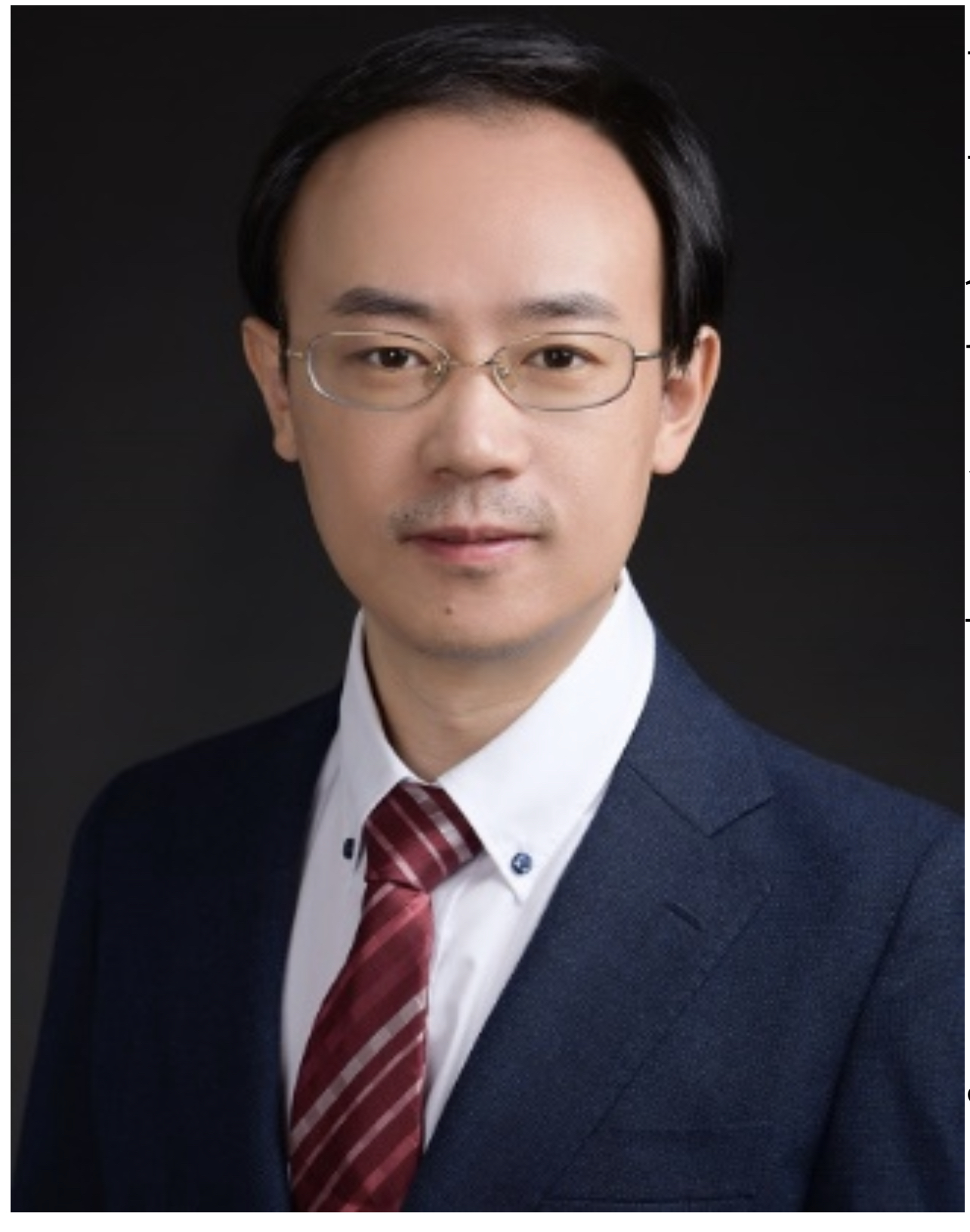}}]{Prof. Xuan Song}  received the Ph.D. degree in signal and information processing from Peking University in 2010. In 2017, he was selected as an Excellent Young Researcher of Japan MEXT. In the past ten years, he led and participated in many important projects as a principal investigator or primary actor in Japan, such as the DIAS/GRENE Grant of MEXT, Japan; Japan/US Big Data and Disaster Project of JST, Japan; Young Scientists Grant and Scientific Research Grant of MEXT, Japan; Research Grant of MLIT, Japan; CORE Project of Microsoft; Grant of JR EAST Company and Hitachi Company, Japan. He served as Associate Editor, Guest Editor, Area Chair, Program Committee Member or reviewer for many famous journals and top-tier conferences, such as IMWUT, IEEE Transactions on Multimedia, WWW Journal, Big Data Journal, ISTC, MIPR, ACM TIST, IEEE TKDE, UbiComp, ICCV, CVPR, ICRA and etc.
\end{IEEEbiography}

\end{document}